\documentclass[letterpaper, 10 pt, conference]{ieeeconf}  % Comment this line out if you need a4paper
\usepackage{amsmath,amsfonts}
\usepackage{array}
\usepackage[caption=false,font=normalsize,labelfont=sf,textfont=sf]{subfig}
\usepackage{textcomp}
\usepackage{stfloats}
\usepackage{url}
\usepackage{verbatim}
\usepackage{graphicx}
\usepackage{cite}
\usepackage{xcolor}

\usepackage{cite}
\usepackage{amsmath,amssymb,amsfonts}
\usepackage{algorithmic}
\usepackage{graphicx}
\usepackage{textcomp}
\usepackage{authblk}
\usepackage{gensymb}

\usepackage{hyperref}
\usepackage{scrextend}
\usepackage{changepage}
\usepackage[margin=0.75in]{geometry}
\usepackage{multirow}
\usepackage{amsfonts}
\usepackage{booktabs}
\usepackage{colortbl}
\usepackage{tikz}
\usepackage{pifont}% http://ctan.org/pkg/pifont
\usepackage[linesnumbered,ruled,vlined]{algorithm2e}

\usepackage[font={footnotesize}]{caption}

\SetCommentSty{mycommfont}

\usepackage{amsmath}
\DeclareMathOperator*{\argmax}{arg\,max}

\IEEEoverridecommandlockouts                              % This command is only needed if 
\title{\LARGE \bf
Transporters with Visual Foresight for \\ Solving Unseen Rearrangement Tasks
}

\author{Hongtao Wu$^{*}$, Jikai Ye$^{*}$, Xin Meng, Chris Paxton, Gregory S. Chirikjian% <-this % stops a space
\thanks{This work is supported by the National Research Foundation, Singapore, under its Medium Sized Centre Programme - Centre for Advanced Robotics Technology Innovation (CARTIN) R-261-521-002-592. Hongtao Wu's PhD tuition is supported by JHU internal funds.}% <-this % stops a space
\thanks{* Equal contribution.}%
\thanks{H. Wu is with The Johns Hopkins University, Baltimore, MD 21218, USA.}
\thanks{J. Ye, X. Meng, and G. S. Chirikjian are with National University of Singapore, Singapore.}
\thanks{C. Paxton is with NVIDIA, USA.}%
\thanks{Address all correspondence to G. S. Chirikjian: mpegre@nus.edu.sg}%        
}

\begin{document}

\maketitle
\thispagestyle{empty}
\pagestyle{empty}

%%%%%%%%%%%%%%%%%%%%%%%%%%%%%%%%%%%%%%%%%%%%%%%%%%%%%%%%%%%%%%%%%%%%%%%%%%%%%%%%
\begin{abstract}
Rearrangement tasks have been identified as a crucial challenge for intelligent robotic manipulation, but few methods allow for precise construction of unseen structures.
We propose a visual foresight model for pick-and-place rearrangement manipulation which is able to learn efficiently.
In addition, we develop a multi-modal action proposal module which builds on the Goal-Conditioned Transporter Network, a state-of-the-art imitation learning method.
Our image-based task planning method, Transporters with Visual Foresight, is able to learn from only a handful of data and generalize to multiple unseen tasks in a zero-shot manner.
TVF is able to improve the performance of a state-of-the-art imitation learning method on unseen tasks in simulation and real robot experiments.
In particular, the average success rate on unseen tasks improves from 55.4\% to 78.5\% in simulation experiments and from 30\% to 63.3\% in real robot experiments when given only tens of expert demonstrations.
Video and code are available on our project website: \url{https://chirikjianlab.github.io/tvf/}
\end{abstract}

%%%%%%%%%%%%%%%%%%%%%%%%%%%%%%%%%%%%%%%%%%%%%%%%%%%%%%%%%%%%%%%%%%%%%%%%%%%%%%%%

\section{INTRODUCTION}
\label{sec: intro}
Prospection enables humans to imagine effects of actions \cite{seligman2013navigating}.
It allows humans to learn multiple tasks from very few examples and generalize to unseen tasks efficiently by combining skills that they have honed in other contexts.
If robots are to become efficient in learning new tasks, this ability will be essential.

One version of prospection is visual foresight, which predicts action effects by hallucinating the expected changes in the robot's observation space. This has been successfully demonstrated in various robotic manipulation tasks~\cite{finn2017deep, paxton2019visual, hoque2020visuospatial, xu2018neural, huang2021dipn, huang2021visual, suh2020surprising}.
However, training an accurate and reliable visual dynamics model requires copious amounts of data.
Previous work~\cite{finn2017deep, hoque2020visuospatial, xu2018neural} collects a large amount of data for training the dynamics model.
In addition, most methods focus on single-task learning with visual foresight~\cite{finn2017deep, paxton2019visual, huang2021dipn, huang2021visual, suh2020surprising}.
While some prior work showcases multi-task learning \cite{hoque2020visuospatial, xu2018neural}, the ability to generalize to unseen tasks which are not present in the training data is still missing.
Moreover, the huge combinatorial search space of possible actions makes planning for complex multi-step tasks computationally challenging.

In this work, we first propose a visual foresight (VF) model which predicts the next-step observation based on the current observation and a pick-and-place action.
Unlike previous methods which encode actions in the latent space \cite{finn2017deep, paxton2019visual}, our model exploits the spatial equivariance in vision-based manipulation tasks by encoding the pick-and-place action in the image space.
This allows our VF model to learn efficiently and predict accurate next-step observations even with only tens of training data.
Secondly, we develop a multi-modal action proposal module by leveraging a state-of-the-art imitation learning method for rearrangement tasks, the Goal-Conditioned Transporter Network (GCTN) \cite{seita2021learning}.
Combining the VF model and the action proposal module with a tree-search algorithm, we propose Transporters with Visual Foresight (TVF), a novel goal-conditioned task planning method for rearrangement tasks that achieves zero-shot generalization to multiple unseen tasks that are structurally similar to the training tasks given only a handful of expert demonstrations.

\begin{figure}[t]
    \centering
    \includegraphics[width=1\columnwidth]{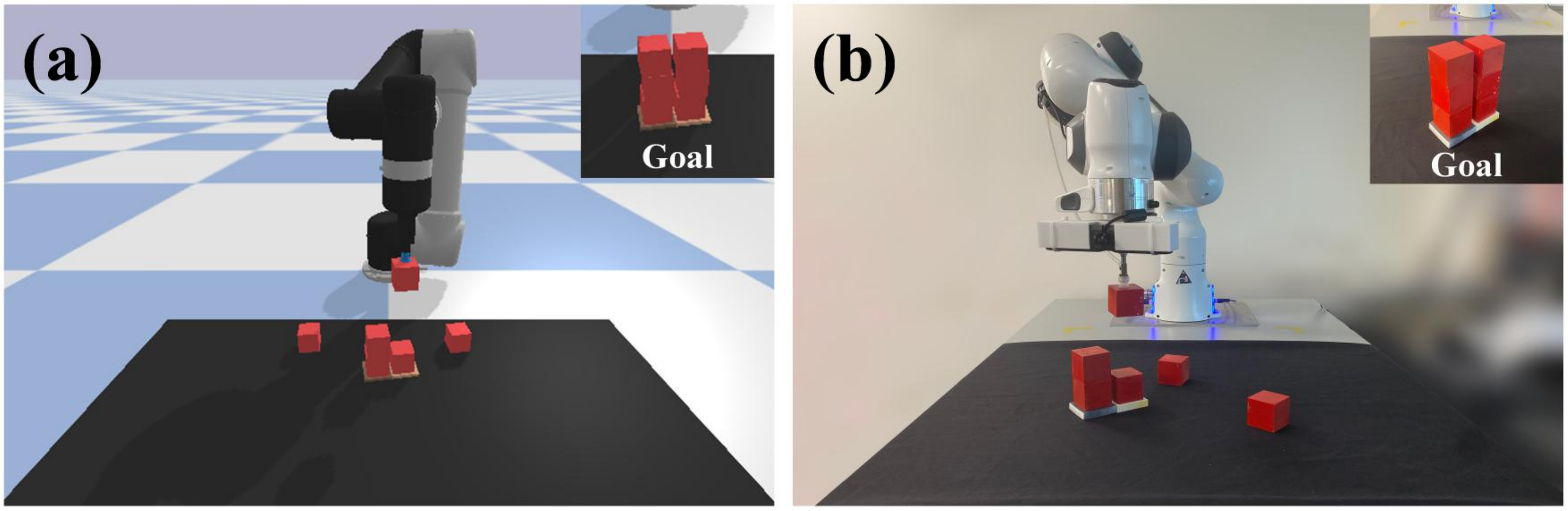}
    \caption{\textbf{Overview}. Given a goal (\textit{e.g.}, a twin tower), the robot plans motions to construct the goal from randomly positioned blocks.}
    \label{fig: overview}
    \vspace{-0.6cm}
\end{figure}

We perform experiments on both simulation and real robot platforms and compare with the state-of-the-art baseline method GCTN \cite{seita2021learning}.
In simulation experiments, our method achieves an average success rate of 78.5\% on 8 unseen tasks given only tens of expert demonstrations, compared to the 55.4\% success rate of GCTN.
In real robot experiments, given 30 expert demonstrations, our method achieves an average success rate of 63.3\% on 3 unseen tasks, outperforming GCTN which achieves 30.0\%.

The key contributions of this work are:
    (1) a goal-conditioned task planning method which achieves zero-shot generalization to unseen, long-horizon rearrangement tasks;
    (2) a visual foresight model which is able to learn efficiently from a handful of data; and 
    (3) a multi-modal action proposal module for more versatile action proposal.

\begin{figure*}
    \centering
    \includegraphics[width=2\columnwidth]{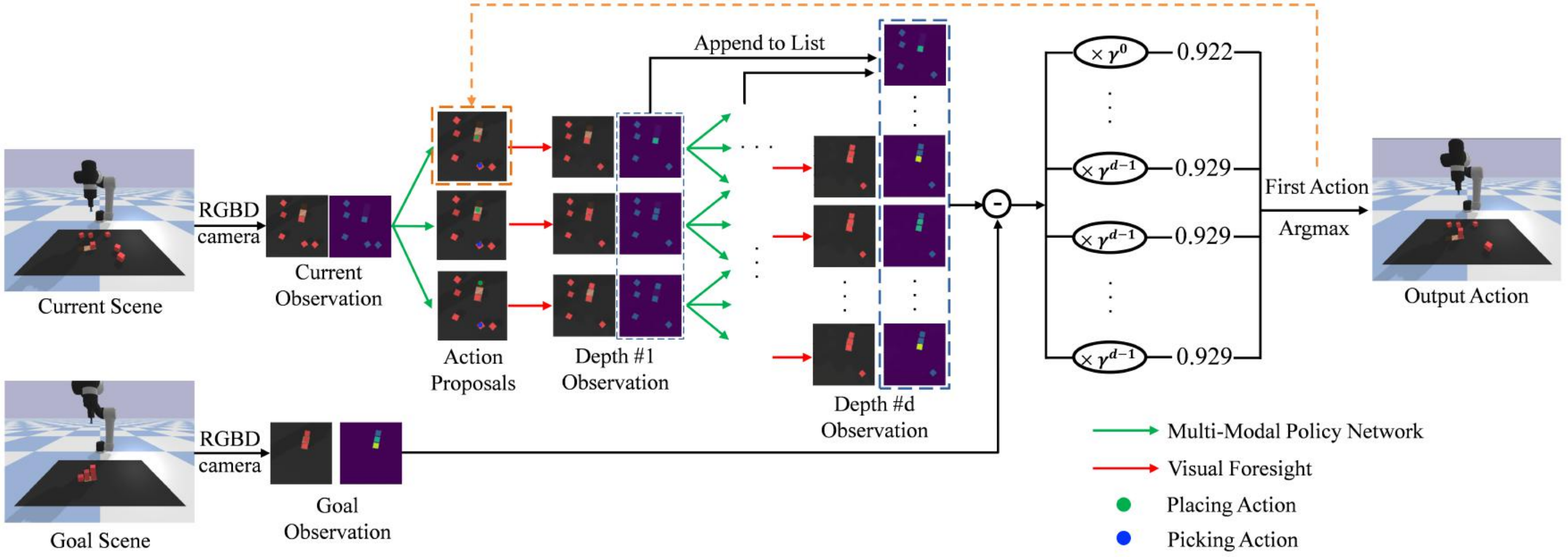}
    \caption{\textbf{Transporters with Visual Foresight (TVF).} Our method takes as input an orthographic top-down view of the current scene and generates multiple actions using a multi-modal action proposal module built on GCTN \cite{seita2021learning}. 
    We then use our proposed visual foresight (VF) model to predict the next-step observation after taking each action.
    Combining the VF model and the action proposal module with a tree-search algorithm, we propose Transporters with Visual Foresight (TVF) for robot rearrangement task planning. 
    See Sec. \ref{sec: methods} for more details.}
    \label{fig: pipeline}
    \vspace{-0.5cm}
\end{figure*}

\section{RELATED WORK}
\label{sec: related_work}
\subsection{Transporter Networks}
The most relevant work is probably the Transporter Networks \cite{zeng2020transporter, seita2021learning, lim2021multi, shridhar2022cliport, huang2022equivariant}.
Zeng \textit{et al.} \cite{zeng2020transporter} propose Transporter Networks (TN) for efficient learning of tabletop rearrangement tasks.
Seita \textit{et al.} \cite{seita2021learning} build on Transporter Networks and propose Goal-Conditioned Transporter Networks (GCTN) for handling deformable objects.
However, both TN and GCTN focus on single-task learning and do not address unseen task generalization.
CLIPort \cite{shridhar2022cliport} extends Transporter Networks to a language-conditioned policy that solves language-specific tabletop tasks.
Lim \textit{et al.} \cite{lim2021multi} propose Sequence-Conditioned Transporter Networks (SCTN) based on GCTN with sequence conditioning for accomplishing a sequence of tasks in a single rollout. 
SCTN uses a human oracle to provide different intermediate next-step goal images for GCTN to be conditioned on throughout a rollout.
Our multi-task setting focuses on a single task for a single rollout instead of a sequence of tasks in a single rollout.
In addition, instead of manually providing multiple intermediate next-step goals during the rollout, in our method, only one single last-step goal is provided and fixed throughout a rollout.

\subsection{Visual Foresight}
Research on visual foresight (VF) has become popular in recent decades \cite{finn2017deep, ebert2017self, paxton2019visual, hoque2020visuospatial, huang2021dipn, kossen2019structured, minderer2019unsupervised}. 
Finn and Levine \cite{finn2017deep} use a convolutional LSTM architecture for video prediction and demonstrate on a robot pushing task with model predictive control (MPC). 
Building on \cite{finn2017deep}, Ebert \textit{et al.} \cite{ebert2017self} account for occlusion by adding temporal skip-connections to the architecture.
In \cite{huang2021dipn}, Huang \textit{et al.} utilize a Mask-RCNN for object segmentation and predict the transformation for each object in the next frame.
Kossen \textit{et al.} \cite{kossen2019structured} discover hidden structures in images using a latent variable model and predict the dynamics in the latent space with a graphical neural network.
Minderer \textit{et al.} \cite{minderer2019unsupervised} also adopt a latent variable model but instead use a keypoint detector for hidden structure extraction.
Paxton \textit{et al.} \cite{paxton2019visual} and Hoque \textit{et al.} \cite{hoque2020visuospatial} train a visual dynamics model with a huge amount of data to infer the next-step image after taking an action and plan based on evaluating the value of the predicted image.
In contrast to requiring copious amounts of data for training, our method learns the visual dynamics model efficiently with only a handful of data.

\subsection{Task Planning}
Task planning has been popular for decades in robotics \cite{toussaint2015logic, garrett2021integrated, kaelbling2010hierarchical, paxton2019visual, hoque2020visuospatial, huang2021visual, song2020multi, mukherjee2021reactive, xu2018neural, mcdonald2022guided, qureshi2021nerp, labbe2020monte, hundt2020good}.
Task and motion planning (TAMP) \cite{toussaint2015logic, garrett2021integrated, kaelbling2010hierarchical} solves complex tasks by integrating discrete high-level planning with continuous low-level planning.
These methods generally rely on full knowledge about the world and state estimation of objects, although learning has been applied to weaken these assumptions recently \cite{mukherjee2021reactive, labbe2020monte}.
Another line of work uses modular perception frameworks to obtain object states \cite{huang2021visual, song2020multi} or latent representations of objects \cite{qureshi2021nerp} and proposes actions with a single-task policy network.
Our method differs from these methods by encoding object states in a visual dynamics model instead of relying on state estimation.

\subsection{Multi-Task and Meta Learning}
There is a growing interest in multi-task learning \cite{caruana1997multitask, mulling2013learning, kalashnikov2021mt, yang2020multi, lim2021multi, hundt2021good} and meta-learning \cite{finn2017one, duan2017one, yu2018one}.
Multi-task learning has been studied in reinforcement learning settings with shared multi-task policies \cite{kalashnikov2021mt, yang2020multi} and gating networks \cite{mulling2013learning}.
Meta-learning \cite{finn2017one, duan2017one, yu2018one} and task-to-task transfer \cite{hundt2021good} aim to learn a policy which is able to perform well on a new task given only a few demonstrations of that task.
In contrast to learning a policy that performs multiple training tasks or a new task given more training data of that task, our goal is to learn a policy that is able to perform multiple unseen tasks that are structurally similar to the training tasks in a zero-shot manner.

\section{PROBLEM FORMULATION}
\label{sec: problem_formulation}
We formulate the planar tabletop rearrangement task as learning a goal-conditioned policy $\pi$ that maps a current observation $\mathbf{o}_t$ and a goal observation $\mathbf{o}_g$ to an action $\mathbf{a}_t$:
\begin{equation}
    \pi (\mathbf{o}_t, \mathbf{o}_g): (\mathbf{o}_t, \mathbf{o}_g) \rightarrow \mathbf{a}_t
\end{equation}
The action is specified by a pair of picking and placing poses $\mathbf{a}_t = (T_{\rm{pick}}, T_{\rm{place}})$ where $T_{\rm{pick}}, T_{\rm{place}} \in SE(2)$.
Similar to \cite{zeng2020transporter}, the observation in our method is an orthographic top-down view $\mathbf{o}_t \in \mathbb{R}^{H\times W\times 4}$ of the tabletop workspace.
Each pixel corresponds to a vertical column in the workspace.
The four channels of $\mathbf{o}_t$ are the RGB channels and the height of the column in the workspace.
We are able to map every pixel $\mathbf{p}=(u,v) \in \mathbf{o}_t$ to a picking (or placing) position on the table via camera-to-robot calibration.
The picking and placing actions are parameterized as $T_{\rm{pick}} = (\mathbf{p}_{\rm{pick}}, \theta_{\rm{pick}})$ and $T_{\rm{place}} = (\mathbf{p}_{\rm{place}}, \theta_{\rm{place}})$.
The goal observation is the top-down view $\mathbf{o}_g \in \mathbb{R}^{H\times W\times 4}$ of the goal scene.
To execute $\mathbf{a}_{t}$, the robot first moves to $T_{\rm{pick}}$ and lowers its end effector.
The gripper is then activated to grasp the object.
After the object is grasped, the robot moves upward and then towards $T_{\rm{place}}$.
After arriving at $T_{\rm{place}}$, the robot lowers its end effector and deactivates the gripper to release the object.

Our goal is to learn a policy that generalizes to multiple unseen tasks in a zero-shot manner.
To train $\pi$, we assume a small dataset containing expert demonstrations of $M$ different tasks $D = \{\xi_{i}\}_{i=1}^{N}$ is provided as training data where $\xi_{i}$ is an episode of a task.
An episode $\xi _{i}$ of length $T_i$ contains observations and actions of different steps:
$\xi_{i} = \{ \mathbf{o}_1, \mathbf{a}_1, \mathbf{o}_2, \mathbf{a}_2, \cdots, \mathbf{o}_{T_i}, \mathbf{a}_{T_{i}}, \mathbf{o}_{T_{i}+1}\}$ where $\mathbf{o}_{T_{i}+1} = \mathbf{o}_{g}$.

\section{METHODS}
\label{sec: methods}
In this section, we first introduce the proposed visual foresight (VF) model.
We then describe the multi-modal action proposal module developed from  Goal-Conditioned Transporter Networks (GCTN)~\cite{seita2021learning}.
Finally, we introduce combining the VF model and the action proposal module with a tree-search algorithm for long-horizon task planning.

\subsection{Visual Foresight (VF) Model}
\label{subsec: method_TVF}
\begin{figure}
    \centering
    \includegraphics[width=0.93\columnwidth]{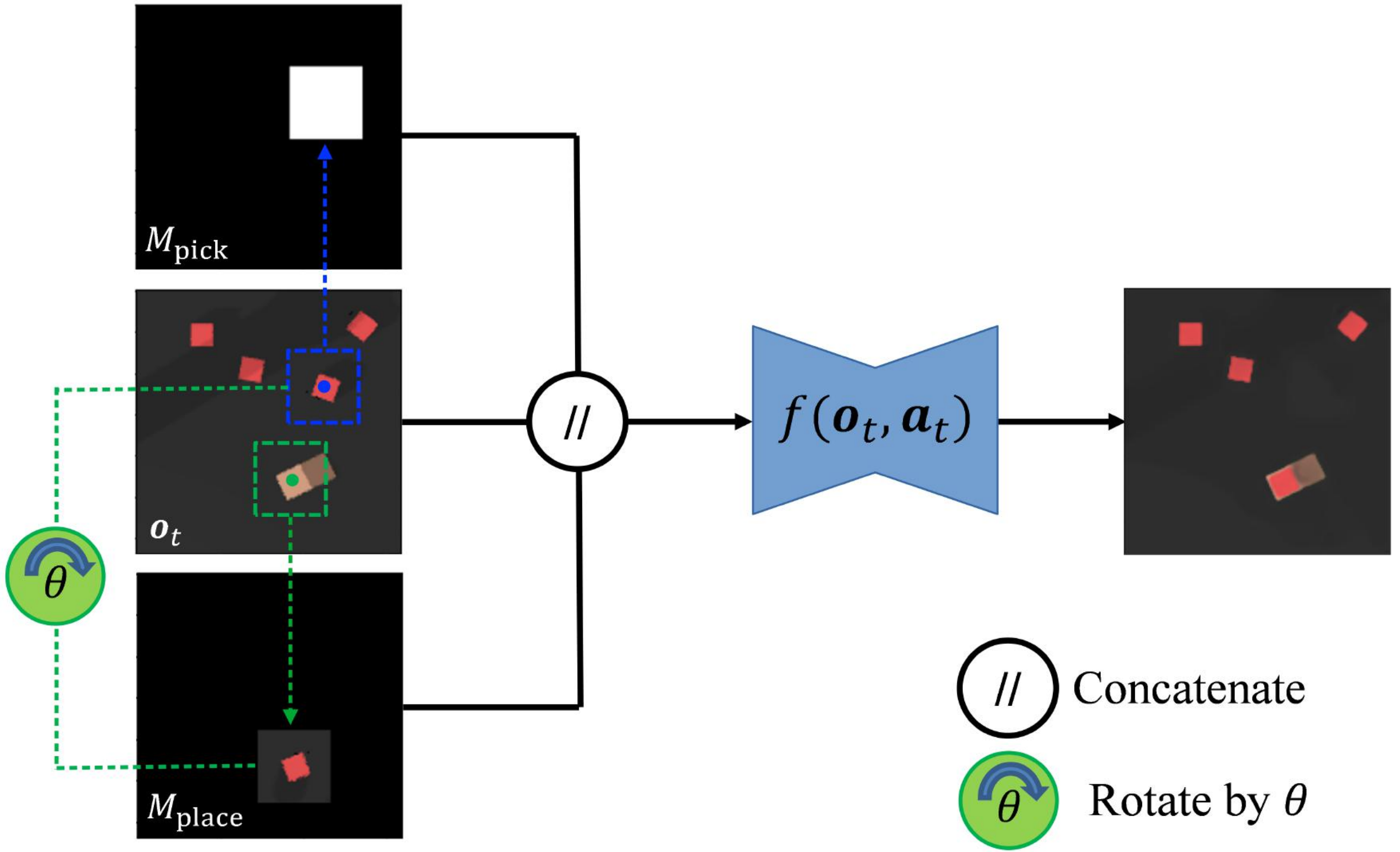}
    \caption{\textbf{Visual Foresight (VF) Model.} Our VF model predicts the next-step observation from the current observation and a pick-and-place action. The blue dot and the green dot indicate $\mathbf{p}_{\rm{pick}}$ and $\mathbf{p}_{\rm{place}}$, respectively. We train the VF model $f(\mathbf{o}_t, \mathbf{a}_t)$ with a small amount of expert demonstrations.}
    \label{fig: tvf}
    \vspace{-0.6cm}
\end{figure}

Fig. \ref{fig: tvf} shows the network architecture of our VF model.
It takes as input the top-down observation ${\mathbf{o}_{t}}$ and a pick-and-place action $\mathbf{a}_t = (T_{\rm{pick}}, T_{\rm{place}})$, and outputs the \textit{imagined} observation of the next step $\mathbf{o}_{t+1} \in \mathbb{R}^{H\times W\times 4}$:
\begin{equation}
    \mathbf{o}_{t+1} = f(\mathbf{o}_t, \mathbf{a}_t)
\end{equation}
$T_{\rm{pick}}$ is specified by a binary image $M_{\rm{pick}}\in \mathbb{R}^{H\times W}$ which is positive within a square mask centered at $\mathbf{p}_{\rm{pick}}$ and zero elsewhere.
To represent $T_{\rm{place}}$, we first rotate $\mathbf{o}_{t}$ by $\Delta \theta =\theta_{\rm{place}}-\theta_{\rm{pick}}$ about a pivot positioned at $\mathbf{p}_{\rm{pick}}$.
The rotated $\mathbf{o}_{t}$ is then cropped at $\mathbf{p}_{\rm{pick}}$ with a square of the same size as the mask in $M_{\rm{pick}}$.
The cropped image is pasted on a zero image at $\mathbf{p}_{\rm{place}}$ to create $M_{\rm{place}} \in \mathbb{R}^{H\times W \times 4}$.
A fully convolutional network (FCN) \cite{long2015fully} takes as input the concatenated image of $\mathbf{o}_{t}$, $M_{\rm{pick}}$, and $M_{\rm{place}}$ and outputs $\mathbf{o}_{t+1}$. 
The FCN is a feed-forward neural network composed of convolution and deconvolution blocks with residual connections \cite{he2016deep}. 
Intuitively, our VF model imagines $\mathbf{o}_{t+1}$ by ``cutting" $\mathbf{o}_t$ with a square mask at $\mathbf{p}_{\rm{pick}}$, rotating the cut by $\Delta \theta$, and ``pasting" the cut by overlaying it on top of $\mathbf{o}_t$ at $\mathbf{p}_{\rm{place}}$.

\textbf{Spatial Equivariance for Efficient Learning.}
Our VF model achieves high sample efficiency by utilizing a spatially consistent input and introducing inductive biases in the network design.
In fact, the dynamics of the tabletop rearrangement problem are $SE(2)$ equivariant; that is, applying a transformation $g \in SE(2)$ on $\mathbf{o}_{t}$ and $\mathbf{a}_{t}$ will lead to an identical transformation on $\mathbf{o}_{t+1}$.
If we define $\mathbf{x}_{t} = (\mathbf{o}_{t}, \mathbf{a}_{t})$, the $SE(2)$ equivariance can be written as:
\begin{equation}
    f(g \bullet \mathbf{x}_{t}) = g \cdot f(\mathbf{x}_{t})
    \label{eqn:equivariance}
\end{equation}
\noindent where $g \bullet \mathbf{x}_{t} \doteq (g \cdot \mathbf{o}_{t}, g \odot \mathbf{a}_{t})$; $\cdot$ indicates the operation on observation $\mathbf{o}_{t}$ which transforms the pixels with $g$; $g \odot \mathbf{a}_{t} \doteq (g\circ T_{\rm{pick}}, g \circ T_{\rm{place}})$ and $\circ$ is the group operation for $SE(2)$.
More details on $SE(2)$ equivariance of the dynamics can be found in the supplementary materials on our project page.
By using a spatially consistent observation and encoding the action in the image space, we are able to take advantage of the $SE(2)$ equivariance and conveniently implement data augmentation by applying the same rigid transform to $\mathbf{o}_t$, $\mathbf{a}_t$, and $\mathbf{o}_{t+1}$, as seen in prior works~\cite{zeng2020transporter,seita2021learning, shridhar2022cliport}.

In addition, encoding actions in the image space instead of the latent space as \cite{finn2017deep, paxton2019visual} allows us to use an FCN as our network architecture and take advantage of the translational equivariance property of the network \cite{long2015fully}.
This property is very desirable and has been previously shown to improve learning efficiency in vision-based manipulation \cite{zeng2020transporter, seita2021learning, shridhar2022cliport}.
While we have achieved translational equivariance with an FCN in this work, the dynamics for tabletop rearrangement is in general $SE(2)$ equivariant.
We leave the investigation on using $SE(2)$ equivariant networks as future work.
In addition, the changes of the scene in tabletop rearrangement mostly happens locally in the vicinity of picking and placing positions.
By using a square mask positioned locally at $\mathbf{p}_{\rm{pick}}$ and $\mathbf{p}_{\rm{place}}$, our VF model intuitively captures this feature -- the FCN only needs to attend to the local regions of picking and placing in $\mathbf{o}_t$.

\begin{figure}
    \centering
    \includegraphics[width=1\columnwidth]{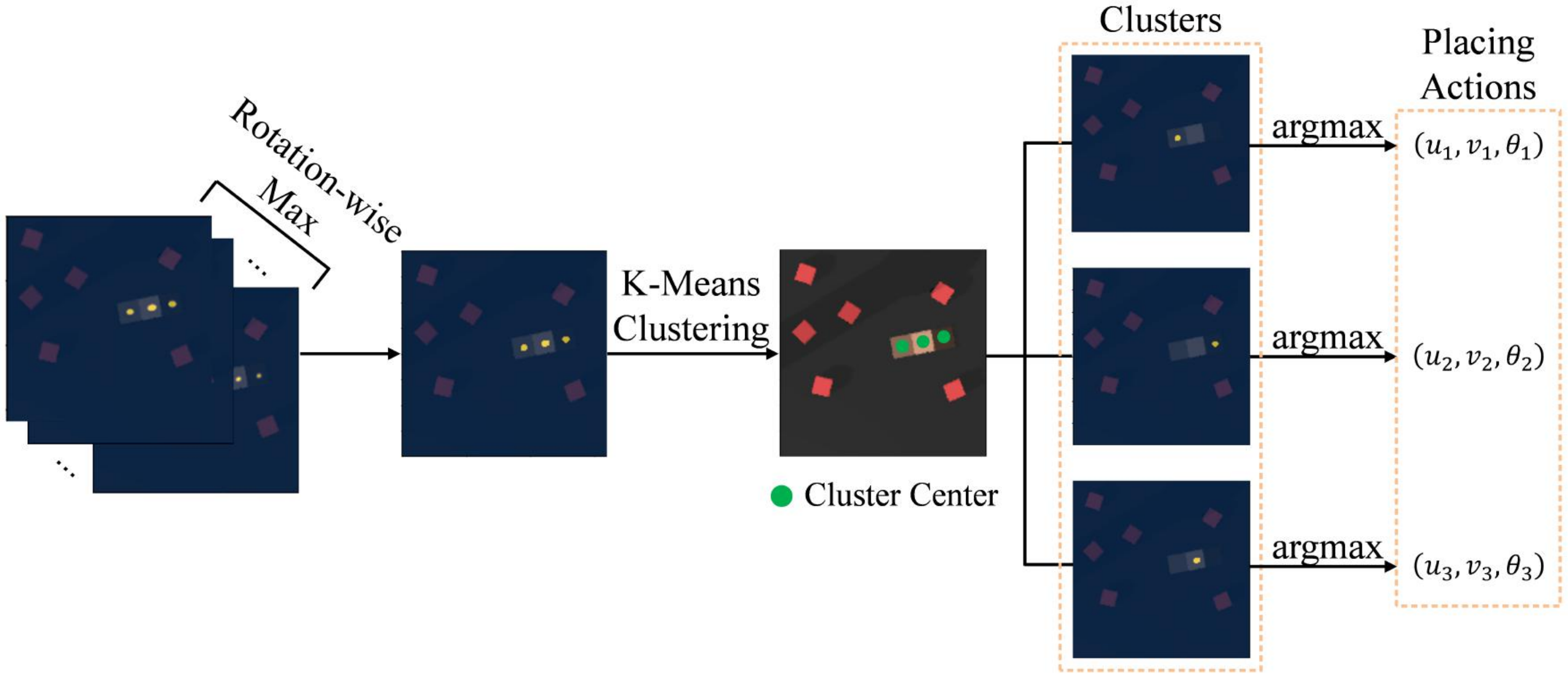}
    \caption{\textbf{Multi-Modal Action Proposal.} 
    We use K-Means Clustering to identify candidate high-level actions over which we can plan.
    The pixels highlighted with yellow indicate placing actions with high-value.
    The green dots in the image after K-Means Clustering are the cluster centers for the three clusters.
    The three maps following show the placing actions for each cluster.}
    \label{fig: kmeans}
    \vspace{-0.6cm}
\end{figure}

\subsection{Multi-Modal Action Proposal}
\label{subsec: method_multi_modal_gctn}
\textbf{Background.}
Goal-Conditioned Transporter Networks (GCTN) \cite{seita2021learning} are a powerful approach for pick-and-place rearrangement manipulation.
Their observation space is also an orthographic top-down view of the tabletop workspace as introduced in Sec.~\ref{sec: problem_formulation}.
GCTN takes as input the observation of the current scene $\mathbf{o}_{t}$ and the goal scene $\mathbf{o}_{g}$ and outputs a pick-and-place action $\mathbf{a}_{t}= (T_{\rm{pick}}, T_{\rm{place}})$.
It consists of 4 FCNs.
The first FCN takes as input $\mathbf{o}_{t}$ and $\mathbf{o}_{g}$ and outputs a dense action-value map $Q_{\rm{pick}} \in \mathbb{R}^{H \times W}$ which correlates with the picking success.
The picking position is given by $\mathbf{p}_{\rm{pick}} = \argmax_{(u, v)}{Q_{\rm{pick}}((u, v)|\mathbf{o}_{t}, \mathbf{o}_{g})}$.
By using a symmetric gripper, \textit{e.g.}, a suction cup gripper, $\theta_{\rm{pick}}$ is set to 0.
For the placing action, the orientation space (\textit{i.e.}, $SO(2)$) is discretized into $R$ bins.
The final three FCNs generate the placing action-value map $Q_{\rm{place}} \in \mathbb{R}^{H \times W \times R}$ which correlates with the placing success.
$Q_{\rm{place}}(u, v, r)$ indicates the placing success of the action $t = (u, v, 2r\pi/R)$.
The placing action is given by $T_{\rm{place}} = \argmax_{t} Q_{\rm{place}}(t | \mathbf{o}_{t}, \mathbf{o}_{g}, T_{\rm{pick}})$.
More details about GCTN can be found in~\cite{seita2021learning}.

\vspace{-0.3cm}
\begin{algorithm}[!ht]
\SetAlgoLined
\DontPrintSemicolon

  $(T_{\rm{pick}}$, $Q_{\rm{place}}) \gets$ GCTN($\mathbf{o}_t$, $\mathbf{o}_g$)\;
  \label{algl:multi-modal:gctn}
  $Q_{\rm{place}}^{\rm{max}} \gets \max_{(u,v, \theta)} {Q}_{\rm{place}}(u,v, \theta)$ \;
  \label{algo:multi-modal:maxQ}
  $\tilde{Q}_{\rm{place}}(u,v) \gets \max_{\theta} Q_{\rm{place}}(u, v, \theta)$\;
  \label{algo:multi-modal:maxrotation}
  $\tilde{\theta}(u, v) \gets \argmax_{\theta} Q_{\rm{place}}(u,v, \theta)$
  
  $S \gets \emptyset$
  
  \ForEach{$(u, v)$}
  {
    \If{$\tilde{Q}_{\rm{place}}(u,v) > \alpha Q_{\rm{place}}^{\rm{max}} $} {append $(u,v)$ to $S$}
  }
    
  $S \gets$ Top\_N($S$)\;
  \label{algo:multi-modal:topn}
  
  $(A_1,A_2,...,A_K) \gets$ K\_Means\_Clustering($S$) \;  %\tcp*{obtain $K$ clusters from $S$}
  \label{algo:multi-modal:kmeans}
  
  \For{$i \gets 1, 2, \cdots K$}
  {
    $(u_i, v_i)\gets \argmax_{(u,v)\in A_i}\tilde{Q}_{\rm{place}}(u,v)$
  
    $\theta_i\gets \tilde{\theta}(u_i,v_i)$  %\tcp*{Retrieve the rotation angle with the largest} $\tilde{Q}$
  
    $T_{\rm{place}}^{i}\gets (u_i,v_i,\theta_i)$
  }
  \Return $T_{\rm{pick}}, T_{\rm{place}}^{1}, T_{\rm{place}}^{2}, ..., T_{\rm{place}}^{K}$
\caption{MultiModalActionProposal($\mathbf{o}_t$, $\mathbf{o}_g$)}
\label{algo: multi-modal}
\end{algorithm}
\vspace{-0.3cm}

\textbf{Multi-Modality with K-means Clustering.}
Instead of outputting only one single action as in GCTN \cite{seita2021learning}, we want to explore more actions for multi-modality and versatility when tackling unseen tasks.
We propose multiple actions by pairing $T_{\rm{pick}}$ generated by maximizing $Q_{\rm{pick}}$ with multiple $T_{\rm{place}}^{i}$ generated from the placing action-value map $Q_{\rm{place}}$ with K-Means Clustering.

The algorithm is given in Alg.~\ref{algo: multi-modal}.
Specifically, given $T_{\rm{pick}}$ and $Q_{\rm{place}}$ outputted by GCTN, we first find the maximum value of $Q_{\rm{place}}$ which we denote as $Q^{\rm{max}}_{\rm{place}}$ (Line \ref{algo:multi-modal:maxQ}).
We then maximize over the rotation channel to get the max rotation map $\tilde{Q}_{\rm{place}} \in \mathbb{R}^{H\times W}$ (Line \ref{algo:multi-modal:maxrotation}).
The pixels with a value larger than $\alpha Q^{\rm{max}}_{\rm{place}}$ are selected and appended to a list $S$, where
$\alpha \in (0, 1)$ is a hyperparameter. 
We use $\alpha$ to filter out pixels of which the values are not substantial -- their corresponding actions are considered bad proposals and will become noise in the clustering.
The top $N$ pixel positions within $S$ are selected (Line \ref{algo:multi-modal:topn}) and used for K-Means Clustering (Line \ref{algo:multi-modal:kmeans}) to generate $K$ clusters.
For each cluster $A_{i}$, the pixel position $(u_i, v_i)$ with the maximum value of $\tilde{Q}_{\rm{place}}$ plus its corresponding rotation angle $\theta_{i}$ is used as the place action $T_{\rm{place}}^{i}$ for the cluster.
See Fig.~\ref{fig: kmeans} for visualization.

\vspace{-0.3cm}
\begin{algorithm}[ht!]
\DontPrintSemicolon
  $n_0 \gets [\mathbf{o}_{t}, 0, \emptyset]$\;
  
  $L \gets \emptyset$
  
  $L_{\rm{prev}} \gets [n_{0}]$
  
  \For{$ i \gets 1, 2, \cdots, d_{\rm{max}}$}
  {
    
    $L_{\rm{curr}} \gets \emptyset$
    
    \ForEach{$n$ \rm{in} $L_{\rm{prev}}$}
    {
        
        $[\mathbf{o}, d, \tau] \gets n$
        \label{algo:treesearch:iteration_start}
        
         $(T_{\rm{pick}}, T_{\rm{place}}^{1}, \cdots , T_{\rm{place}}^{K}) \gets$ MultiModalActionProposal$ (\mathbf{o}, \mathbf{o}_g)$ \;
        \label{algo:treesearch:actionproposal}
    
        \For{$ k \gets 1, 2, \cdots, K$}
        {
            $\mathbf{o}' \gets$ $f(\mathbf{o}, (T_{\rm{pick}}, T_{\rm{place}}^{k}))$\;
            \label{algo:treesearch:tvf}
        
            $\tau' \gets \tau \cup \{(T_{\rm{pick}}, T_{\rm{place}}^{k})\}$\;
            \label{algo:treesearch:trajectory}
        
            $n' \gets [\mathbf{o}', d + 1, \tau']$
            \label{algo:treesearch:node}
            
            $L_{\rm{curr}} \gets L_{\rm{curr}} \cup \{n'\}$\;
            \label{algo:treesearch:iteration_end}
        }
    }
    
    $L_{\rm{prev}} \gets L_{\rm{curr}}$
    
    $L \gets L \cup L_{\rm{curr}}$
  }
  
  \Return{$L$}
  
\caption{TreeSearch($\mathbf{o}_{t}$, $\mathbf{o}_{g}$, $d_{\rm{max}}$)}
\label{algo:treesearch}
\end{algorithm}
\vspace{-0.4cm}

\subsection{Transporters with Visual Foresight}
\label{subsec: method_tree_search}
Combining the VF model (Sec. \ref{subsec: method_TVF}) and the multi-modal action proposal module (Sec. \ref{subsec: method_multi_modal_gctn}) with a full tree-search algorithm, we propose Transporters with Visual Foresight (TVF). % , a task planning method.
Alg.~\ref{algo:treesearch} shows the tree search algorithm.
The maximum depth $d_{\rm{max}}$ of the tree is a hyperparameter. 
Each edge in the tree corresponds to an action.
Each node $n$ contains the current observation $\mathbf{o}$, the depth of the node $d$, and the action sequence $\tau$ which leads the root node $n_{0}$ to the current node.
A typical tree search iteration (Line \ref{algo:treesearch:iteration_start}-\ref{algo:treesearch:iteration_end}) consists of 2 steps: action proposal and node expansion. 
The action proposal takes as input the observation of the node $\mathbf{o}$ and the goal $\mathbf{o}_{g}$ and outputs a picking action and multiple companion placing actions (Line \ref{algo:treesearch:actionproposal}).  
The VF model expands the node by taking each pick-and-place action pair to generate the imagined observation of the next step $\mathbf{o}'$ (Line \ref{algo:treesearch:tvf}).
$\mathbf{o}'$ is used to construct a new node for a new search iteration (Line \ref{algo:treesearch:node}).
Throughout the expansion, we maintain a list $L$ which contains all nodes in the tree.

Alg.~\ref{algo:tvf-gctn} shows the algorithm of TVF.
After the tree is fully expanded, the value of each node is given by $C - L_1(\mathbf{o},\mathbf{o}_{g})$ in which $C$ is a positive constant; $L_1(\cdot, \cdot)$ is the mean absolute error; $\mathbf{o}$ and $\mathbf{o}_g$ are the imagined observation of the node and the goal observation, respectively (Line \ref{algo:tvf-gctn:value}). 
To bias the policy towards short search paths, the value is further multiplied by a discount factor $\gamma^{d-1}$ which decays with the increase of the depth $d$ as $\gamma \in (0, 1)$.
The node with the largest value is chosen.
The robot takes the first action in the action sequence of the node (Line \ref{algo:tvf-gctn:first}), and then replans until the task is accomplished or the step number exceeds the maximum step number.

\vspace{-0.3cm}
\begin{algorithm}
\DontPrintSemicolon
  
  $L \gets$ TreeSearch($\mathbf{o}_{t}$, $\mathbf{o}_{g}$, $d_{\rm{max}}$)
  
  $v_{\rm{max}} = 0$
  
  \ForEach{$n$ \rm{in} $L$}
  {
    $[\mathbf{o}, d, \tau] \gets n$
    
    $v_{\rm{tmp}} \gets \gamma ^{d-1} (C - L_1(\mathbf{o}, \mathbf{o}_g))$ \;
    \label{algo:tvf-gctn:value}
    
    \If{$v_{\rm{tmp}} > v_{\rm{max}}$}
    {
        $v_{\rm{max}} \gets v_{\rm{tmp}}$
        
        $n_{\rm{best}} \gets n$
    }
  }
  
  $(T_{\rm{pick}}$, $T_{\rm{place}}) \gets$ FirstAction($n_{\rm{best}}$)
  \label{algo:tvf-gctn:first}
  
  return $T_{\rm{pick}}$, $T_{\rm{place}}$
  
\caption{TVF($\mathbf{o}_{t}$, $\mathbf{o}_{g}$, $d_{\rm{max}}$)}
\label{algo:tvf-gctn}
\end{algorithm}

\vspace{-0.5cm}

\begin{figure*}
    \centering
    \includegraphics[width=1.85\columnwidth]{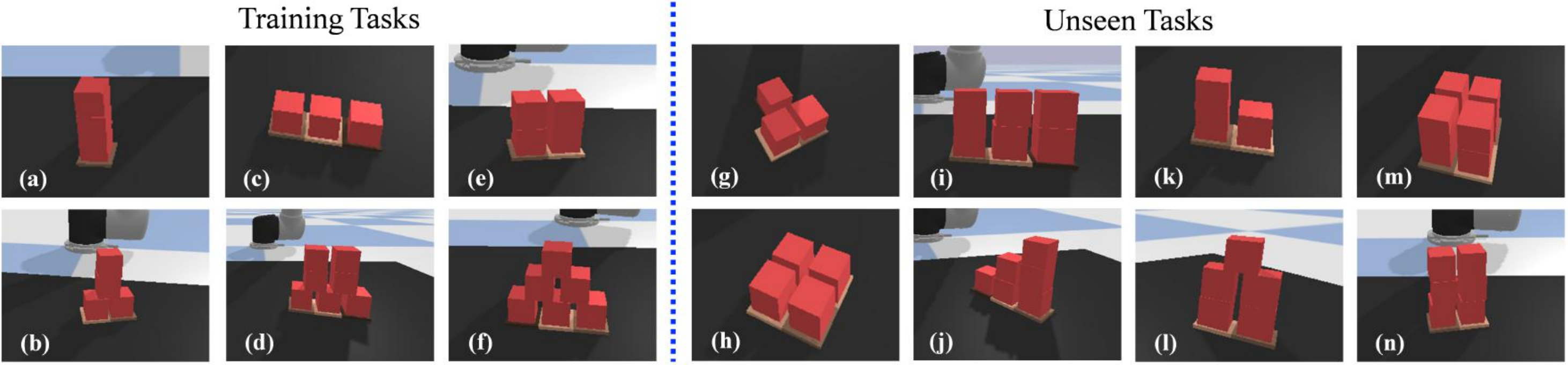}
    \caption{\textbf{Tasks.} (a) Tower. (b) Inverse T-shape. (c) Row. (d) Palace. (e) Square. (f) Pyramid. (g) Plane T. (h) Plane Square. (i) Rectangle. (j) Stair 3. (k) Stair 2. (l) Building. (m) Pallet. (n) Twin Tower. The first 6 tasks are used for training and the rest 8 tasks are unseen tasks which are not present in the training data.}
    \label{fig: tasks}
    \vspace{-0.7cm}
\end{figure*}

\section{EXPERIMENTS}
\label{sec: experiments}
In this section, we answer three questions:
(1) Is TVF able to generalize to unseen tasks in a zero-shot manner after training on a handful of expert demonstrations?
(2) Does TVF work on real robot platforms?
(3) Does the proposed VF model produce better prediction results than baseline methods?
Ablation studies and more experiment details can be found in the supplementary materials on the project website.

\subsection{Simulation Experiments}
Fig. \ref{fig: overview}(a) shows our simulation experiment setup which builds on an open-source manipulation task simulation environment Ravens \cite{zeng2020transporter}. 
A UR5 robot arm with a suction gripper is used to perform pick-and-place actions in a $0.5\times 0.5\rm{m}^{2}$ tabletop workspace.
Three RGB-D cameras are used to reconstruct the top-down observation $\mathbf{o}_{t}$ of the workspace.
We design 14 different block rearrangement tasks (Fig. \ref{fig: tasks}).
In each task, the blocks are randomly positioned and oriented within the workspace.
All tasks require multiple steps to finish.
All tasks are multi-modal -- there may be multiple valid actions to perform in a step.
A scripted oracle is written for each task to provide expert demonstrations leading to the goal configuration.

We divide the 14 tasks into 6 training tasks and 8 unseen tasks which are not present in the training data (Fig. \ref{fig: tasks}).
For each training task, we collect 1000 expert demonstrations.
When testing the performance on training with different numbers of demonstrations (Sec. \ref{subsec:experiments:results:unseen}), we sample from these demonstrations as training data.
To make the VF model more flexible, two random actions, which pick a block on the tabletop and place it at a collision-free pose, are also included in the collection of each expert demonstration.
Both random actions and oracle actions are used for training the VF model; only oracle actions are used for training the GCTN for the multi-modal action proposal.
For all the unseen tasks, we collect 20 demonstrations for testing.

We compare our method with GCTN \cite{seita2021learning}, a state-of-the-art method for learning robot rearrangement tasks.
We also compare different variants of TVF: we vary the number of clusters $K$ in K-Means Clustering and the maximum depth $d_{\rm{max}}$ of the tree.
The cluster numbers $K$ for TVF-Small and -Large are 2 and 3, respectively. 
The tree depths $d_{\rm{max}}$ are 1 and 3, respectively.
Each step takes about 0.08s, 0.14s, and 1.82s for GCTN, TVF-Small, and TVF-Large on an NVIDIA RTX 3090 GPU, respectively.
We evaluate the performance of a method with the success rate.
The maximum step number for a rollout equals the number of blocks in the task.
A trial is considered successful if the planar translation, the z-coordinate, and the rotation about the z-axis of \textit{all} the blocks to the corresponding target poses are less than 1cm, 0.5cm, and 15\degree, respectively.
We train GCTN and TVF variants with 1, 10, 100, and 1000 demonstrations per training task (6, 60, 600, 6000 demonstrations in total).
Following \cite{seita2021learning}, we use different TensorFlow seeds to train 3 models for all methods.
We test each model on the test data and report the average result of the 3 models.

\subsection{Real Robot Experiments} 
Fig. \ref{fig: overview}(b) shows the setup of our real robot experiments.
We implement our method on a Franka Emika Panda robot arm with a suction gripper. 
A PrimeSense Carmine 1.09 RGB-D camera is mounted on the end-effector of the robot.
As in the simulation, the workspace is a $0.5\times 0.5\rm{m}^{2}$ tabletop.
At the beginning of each task, the goal block configuration is shown to the robot and the blocks are then dissembled and placed randomly on the table.
At the beginning of each step, the robot goes to a predefined configuration to capture the top-down observation of the current scene.
We use the height from the observation of the picking and placing locations to determine the height of the picking and placing actions.
IKFast \cite{diankov2010automated} and MoveIt \cite{chitta2012moveit} are used for motion planning.
We pre-process the RGB images and heightmaps by filtering out the background.
We found that GCTN is not able to learn well without the background filtering.
More details can be found in the supplementary materials on the project website.

\subsection{Training Details}
We implement our method with TensorFlow.
Taking advantage of the spatial consistency of our method (Sec. \ref{subsec: method_TVF}), we apply extensive data augmentation of random translations and rotations to the training data to train the VF model.
We use the L1 loss and the Adam optimizer with a learning rate of $1\times 10^{-4}$ and train for 60000 iterations.
For the L1 loss, the weight for the height channel is five times larger than those for the RGB channels.
To train GCTN, we use exactly the same training setting as \cite{seita2021learning}. 
Both simulation and real robot experiments are trained with the same setting as described above.

\begin{table}[t]
  \setlength\tabcolsep{9pt}
  \caption{
  \textbf{Average Success Rates of Simulation Experiments on Unseen Tasks.}
  We show the average success rate (\%) on the test data of unseen tasks v.s. \# of demonstrations (1, 10, 100, or 1000) per task in the training data.
  Higher is better.
  }
  \centering
  \footnotesize
  \begin{tabular}{@{}lccccccccc@{}} % centering
  \toprule
 Method & 1 & 10 & 100 & 1000\\
 \midrule
 GCTN                                &          1.3 &          55.4 &          49.0 &          54.2\\
 TVF-Small ($K=2$, $d_{\rm{max}}=1$) &          1.7 &          71.5 &          62.3 &          72.5\\
 TVF-Large ($K=3$, $d_{\rm{max}}=3$) & \textbf{2.9} & \textbf{78.5} & \textbf{71.7} & \textbf{85.6}\\
  \bottomrule \vspace{-0.8em} \\
  \end{tabular}
  \label{tab:results:average}
  \vspace{-0.5cm}
\end{table}

\subsection{Result I: Simulation Experiments}
\label{subsec:experiments:results:unseen}
We evaluate on zero-shot generalization to unseen tasks which are not present in the training data.
We use the models trained with the data of the 6 training tasks and test on the 8 unseen tasks in Fig. \ref{fig: tasks} without additional training.
The average success rates are shown in Tab. \ref{tab:results:average}.
Both TVF variants outperform GCTN on unseen tasks in all cases of demo numbers.
When there are more than 1 demo for each training task, TVF variants outperform GCTN by a large margin. 
Remarkably, TVF-Large achieves an average success rate of 78.5\% when only 10 demos per task is given for training, while vanilla GCTN achieves 55.4\%.
In the case of 1000 demos per task, TVF-Large achieves a remarkable average success rate of 85.6\%.
For the TVF variants, the performance improves with the increase of the complexity of the tree.
The success rates of GCTN and TVF variants are very low when there is only 1 demo per training task and increase by a large margin when given 10 demos per task.
In general, the performance of GCTN and TVF variants increases with the increase of demo numbers.
However, this is not always the case: the performance of 100 demos is worse than that of 10 demos.
The original GCTN paper reports similar results \cite{seita2021learning}.

The success rates for each unseen task are shown in Tab. \ref{tab:results:unseen_tasks}.
In the cases of 10, 100, and 1000 demos, both TVF variants outperform GCTN in every task.
For Building with 1000 demos, GCTN achieves an average success rate of 3.3\% while TVF-Large improves the performance up to 25.0\%.
Both GCTN and TVF variants are able to achieve good results on simple tasks (\textit{e.g.}, Plane Square and Plane T).
But for more complicated tasks, GCTN struggles while TVF variants perform much better.
Another interesting observation is that the performance of TVF variants correlates with that of GCTN.
This is because the action proposal module is based on GCTN -- if GCTN is not good, the multi-modal action proposal will not be good either.

\begin{table}[t]
  \setlength\tabcolsep{4.5pt}
  \caption{
  \textbf{Success Rates of Simulation Experiments on Unseen Tasks.}
  We show the success rate (\%) on the test data of each unseen task v.s. \# of demonstrations (1, 10, 100, or 1000) per task in the training data.
  * indicates that the success rates of \textit{both} TVF variants are at least 20\% higher than that of GCTN.
  Higher is better.
  }
  \centering
  \footnotesize
  \begin{tabular}{@{}lccccccccc@{}} % centering
  \toprule
 & \multicolumn{4}{c}{Plane Square} & \multicolumn{4}{c}{Plane T} \\
 \cmidrule(lr){2-5} \cmidrule(lr){6-9}
 Method & 1 & 10 & 100 & 1000 & 1 & 10 & 100 & 1000\\
 \midrule
  GCTN      &          1.7 &           86.7 &          95.0 &           96.7   &           5.0 &          78.3 &           93.3 &          90.0\\
  TVF-Small &          3.3 &           98.3 & \textbf{96.7} & \textbf{100.0}   &           3.3 & \textbf{90.0} & \textbf{100.0} & \textbf{98.3}\\
  TVF-Large & \textbf{5.0} & \textbf{100.0} & \textbf{96.7} &           98.3   & \textbf{15.0} &          86.7 & \textbf{100.0} &          95.0\\
  \midrule
  & \multicolumn{4}{c}{Stair 2} & \multicolumn{4}{c}{Twin Tower}\\
 \cmidrule(lr){2-5} \cmidrule(lr){6-9} 
 Method & 1 & 10 & 100* & 1000* & 1 & 10 & 100* & 1000\\
 \midrule
  GCTN      &          3.3 &          85.0 &          46.7 &           68.3   & 0.0 &          88.3 &          55.0 &          85.0\\
  TVF-Small & \textbf{6.7} & \textbf{98.3} &          71.7 &           90.0   & 0.0 & \textbf{98.3} & \textbf{85.0} & \textbf{93.3}\\
  TVF-Large &          3.3 &          96.7 & \textbf{95.0} & \textbf{100.0}   & 0.0 &          96.7 & \textbf{85.0} &          91.7\\
 \midrule
 & \multicolumn{4}{c}{Stair 3} & \multicolumn{4}{c}{Building}\\
 \cmidrule(lr){2-5} \cmidrule(lr){6-9} 
 Method & 1 & 10 & 100 & 1000* & 1 & 10 & 100 & 1000\\
 \midrule
  GCTN      & 0.0 &          45.0 &          23.3 &          16.7   & 0.0 &           5.0 &           0.0 &           3.3\\
  TVF-Small & 0.0 &          63.3 &          33.3 &          46.7   & 0.0 &           8.3 & \textbf{10.0} &          11.7\\
  TVF-Large & 0.0 & \textbf{81.7} & \textbf{56.7} & \textbf{90.0}   & 0.0 & \textbf{13.3} &           6.7 & \textbf{25.0}\\
 \midrule
  & \multicolumn{4}{c}{Pallet} & \multicolumn{4}{c}{Rectangle}\\
 \cmidrule(lr){2-5} \cmidrule(lr){6-9} 
 Method & 1 & 10* & 100 & 1000* & 1 & 10* & 100 & 1000*\\
 \midrule
  GCTN      & 0.0 &          23.3 &          51.7 &          31.7   & 0.0 &          31.7 &          26.7 &          41.7\\
  TVF-Small & 0.0 &          60.0 &          61.7 &          65.0   & 0.0 &          55.0 &          40.0 &          75.0\\
  TVF-Large & 0.0 & \textbf{75.0} & \textbf{70.0} & \textbf{90.0}   & 0.0 & \textbf{78.3} & \textbf{63.3} & \textbf{95.0}\\
  \bottomrule \vspace{-0.8em} \\
  \end{tabular}
  \label{tab:results:unseen_tasks}
\end{table}

\begin{figure*}
    \centering
    \includegraphics[width=2\columnwidth]{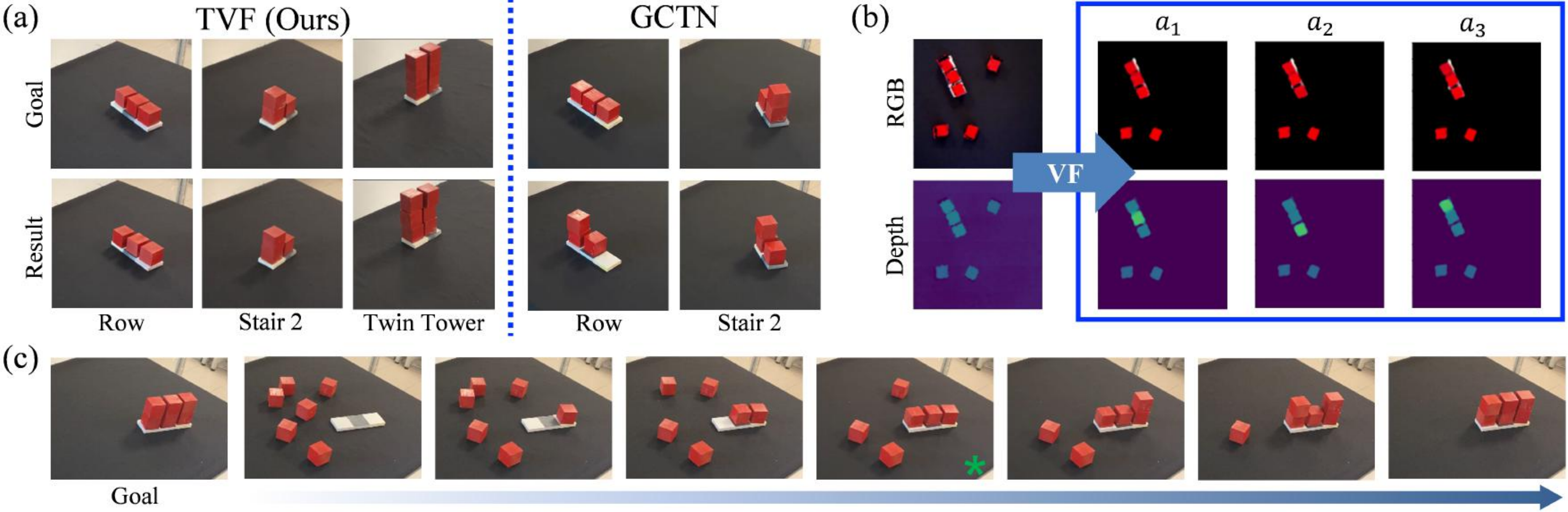}
    \caption{\textbf{Real Robot Experiment Results.} (a) shows the results of TVF and some failure cases of GCTN. (b) shows the visual foresight prediction of the next-step observation after taking three different actions proposed by the multi-modal action proposal module. The current observation on the left is captured from the scene with a green star (*) in (c). Notice that the three actions are different, placing the picked block on the top of the three blocks on the base, respectively. (c) shows a full rollout of TVF on the Rectangle task. The leftmost image shows the goal. The rest images show different steps of the rollout.}
    \label{fig: real}
    \vspace{-0.5cm}
\end{figure*}

\subsection{Result II: Real Robot Experiments}
\label{subsec:experiments:real}
As we are able to train TVF with only a handful of expert data, this makes real robot experiments possible.
We collect 30 expert demonstrations for 3 training tasks (10 demos per task) and use these data to train TVF and GCTN.
We test TVF and GCTN on both the 3 training tasks and 3 unseen tasks.
Each task is tested with 10 rollouts.
The results are shown in Tab. \ref{tab:results:real} and Fig. \ref{fig: real}.
Our method works in the real world.
And it outperforms GCTN in 5 of the 6 tasks and is on par with GCTN on Rectangle.
Notably, while GCTN fails in all 10 rollouts in the unseen Twin Tower task, our method is able to achieve a success rate of 60\%. 

\begin{table}[t]
  \setlength\tabcolsep{4pt}
  \caption{
  \textbf{Real Robot Experiment Success Rate.} 
  We show the success rate (\%) on the test data of all the tasks in the real robot experiments.
  For TVF, we use the TVF-Small variant but increase $K$ to 3 for more action modality.
  * indicates that the success rate of TVF is at least 20\% higher than that of GCTN.
  Higher is better.
  }
  \centering
  \footnotesize
  \begin{tabular}{@{}lccccccccc@{}} % centering
  \toprule
  & \multicolumn{3}{c}{Training Tasks} & \multicolumn{3}{c}{Unseen Tasks} \\
 \cmidrule(lr){2-4} \cmidrule(lr){5-7}
 Method & Row* & Tower & Square* & Stair 2* & Rectangle & Twin Tower*\\
 \midrule
 GCTN           & 60.0  & 90.0 & 70.0 & 40.0 & \textbf{50.0} & 0.0 \\
 TVF       & \textbf{100.0} & \textbf{100.0} & \textbf{90.0} & \textbf{80.0} & \textbf{50.0} & \textbf{60.0}\\
  \bottomrule \vspace{-0.8em} \\
  \end{tabular}
  \label{tab:results:real}
  \vspace{-0.5cm}
\end{table}

\subsection{Result III: Visual Foresight Model}
\label{subsec:experiments:results:dynamics}
Finally, we evaluate our VF model on predicting the next-step observation $\mathbf{o}_{t+1}$ from the current observation $\mathbf{o}_{t}$ and action $\mathbf{a}_{t}$ given a small number of training data.
We compare with a baseline method, referred to as \textit{Latent Dynamics}, which instead encodes the action in the latent space similar to \cite{finn2017deep, paxton2019visual}.
Latent Dynamics also takes as input $\mathbf{o}_{t}$ and $\mathbf{a}_{t}$ and outputs $\mathbf{o}_{t+1}$.
An encoder is first used to encode $\mathbf{o}_{t}$ to a latent representation $L \in \mathbb{R}^{40\times40\times64}$.
The pick-and-place action parameters (\textit{i.e.}, $u_{\rm{pick}}, v_{\rm{pick}}, u_{\rm{place}}, v_{\rm{place}}, \sin(\theta_{\rm{place}}), \cos(\theta_{\rm{place}})$) are tiled into a $40\times40\times6$ map and concatenated channel-wise with $L$ to get $L' \in \mathbb{R}^{40\times40\times70}$.
The next-step observation $\mathbf{o}_{t+1}$ is reconstructed from $L'$ with a decoder.
The encoder-decoder network architecture is an FCN similar to that in our VF model.

We train both methods with 10 demos per training tasks in the simulation (60 demos in total).
For both methods, we use different TensorFlow seeds to train 3 models and report the average result of the 3 models on test data in Tab. \ref{tab:dynamics_result}.
Our VF model outperforms Latent Dynamics in both training and unseen tasks.

\begin{table}[t]
    \setlength\tabcolsep{10pt}
    \caption{
    \textbf{Results on Visual Foresight Models.} 
    The table shows the results of different methods trained with 10 demos for each training task (60 demos in total). 
    It shows the L1 loss of the RGB color channels and the height channel between the predicted observation and the ground truth observation.
    The images are normalized.
    The actions are the expert actions in the expert demonstrations.
    Lower is better.}
    \centering
    \begin{tabular}{@{}lccccc@{}}
    \toprule
        & \multicolumn{2}{c}{Training Tasks} & \multicolumn{2}{c}{Unseen Tasks}\\
        \cmidrule(lr){2-3} \cmidrule(lr){4-5} 
        Method & Color & Height & Color & Height\\
        \midrule
        Latent Dynamics   & 0.0875 & 0.0873 & 0.1441 & 0.1505 \\
        Ours      & \textbf{0.0242} & \textbf{0.0136} & \textbf{0.0691} & \textbf{0.0380} \\
    \bottomrule
    \end{tabular}
    \label{tab:dynamics_result}
    \vspace{-0.5cm}
\end{table}

\section{DISCUSSIONS \& FUTURE WORK}
\label{sec: discussions}
A typical reason for the failures of GCTN is the single-modal action proposal.
In both training tasks and unseen tasks in real robot experiments (Fig. \ref{fig: real}(a)), we observe that the action with the highest value of GCTN is sometimes incorrect.
And since GCTN is single-modal, it will take the incorrect action and fail the task.
On the other hand, even if the action with the highest value is incorrect, TVF is able to predict which actions in the multiple proposals will better lead to the goal with the VF model and take the action.
This brings about substantial advantages on unseen tasks compared to GCTN.
See the columns highlighted with a * in Tab. \ref{tab:results:unseen_tasks} and \ref{tab:results:real}.
Our VF model features efficient training which is able to predict accurate next-step observations given only tens of training data.
This is achieved by introducing inductive biases in the network design -- encoding the action in the image space in a spatially consistent way allows our VF model to take advantage of the translational equivariance property of the FCN.
The performance advantage over Latent Dynamics in Tab. \ref{tab:dynamics_result} justifies our design choice.

As TVF assumes no prior knowledge of objects, we envision it can be generalized to more variable objects.
Moreover, although we focus on $SE(2)$ tabletop rearrangement in this paper and use FCNs as the network architecture, future work can explore applying a more advanced architecture \cite{weiler2019general} that preserves $SE(2)$ equivariance to further improve sample efficiency. 
Another possible direction is to extend the idea of geometry-aware visual foresight to more general settings (\textit{e.g.}, 3D workspaces described by point cloud data \cite{thomas2018tensor}) and develop VF models for accomplishing more challenging tasks.
In addition, our VF model is deterministic, \textit{i.e.}, there is no stochasticity to address the uncertainty of robot actions.
Future work can also explore using uncertainty-aware models (\textit{e.g.}, Bayesian Neural Network \cite{blundell2015weight}) to model the visual dynamics.
Finally, the number of clusters $K$ is fixed for a particular TVF variant, \textit{i.e.}, the number of actions proposed by a particular TVF variant is fixed and equals $K$. 
Future work can investigate methods to adaptively generate different numbers of actions according to the action-value maps $Q_{\rm{pick}}$ and $Q_{\rm{place}}$.

\section{CONCLUSIONS}
\label{sec: conclusions}
In this paper, we propose a simple visual foresight (VF) model which is able to predict the next-step observation from the current observation and a pick-and-place action.
The VF model is able to learn efficiently from only a handful of training data.
In addition, we propose a multi-modal action proposal module which builds on a state-of-the-art imitation learning method \cite{seita2021learning} for more versatile action proposal.
Combining the VF model and the action proposal module with a tree-search algorithm, we propose Transporters with Visual Foresight (TVF), a novel method for rearrangement task planning from image data which is able to achieve zero-shot generalization to unseen tasks with only tens of expert demonstrations.
Results show that our method outperforms a state-of-the-art baseline method on average success rates of unseen tasks in both simulation and real robot experiments.
Our proposed VF model outperforms a baseline method when only given a small number of training data.
Robotic systems that can generalize their function to scenarios beyond interpolations of those that are previously seen represents a higher level of capability.

%%%%%%%%%%%%%%%%%%%%%%%%%%%%%%%%%%%%%%%%%%%%%%%%%%%%%%%%%%%%%%%%%%%%%%%%%%%%%%%%

%%%%%%%%%%%%%%%%%%%%%%%%%%%%%%%%%%%%%%%%%%%%%%%%%%%%%%%%%%%%%%%%%%%%%%%%%%%%%%%%

%%%%%%%%%%%%%%%%%%%%%%%%%%%%%%%%%%%%%%%%%%%%%%%%%%%%%%%%%%%%%%%%%%%%%%%%%%%%%%%%

% \section*{ACKNOWLEDGMENT}

%%%%%%%%%%%%%%%%%%%%%%%%%%%%%%%%%%%%%%%%%%%%%%%%%%%%%%%%%%%%%%%%%%%%%%%%%%%%%%%%
\bibliographystyle{IEEEtran}
\bibliography{reference.bib}

\begin{thebibliography}{10}
\providecommand{\url}[1]{#1}
\csname url@samestyle\endcsname
\providecommand{\newblock}{\relax}
\providecommand{\bibinfo}[2]{#2}
\providecommand{\BIBentrySTDinterwordspacing}{\spaceskip=0pt\relax}
\providecommand{\BIBentryALTinterwordstretchfactor}{4}
\providecommand{\BIBentryALTinterwordspacing}{\spaceskip=\fontdimen2\font plus
\BIBentryALTinterwordstretchfactor\fontdimen3\font minus
  \fontdimen4\font\relax}
\providecommand{\BIBforeignlanguage}[2]{{%
\expandafter\ifx\csname l@#1\endcsname\relax
\typeout{** WARNING: IEEEtran.bst: No hyphenation pattern has been}%
\typeout{** loaded for the language `#1'. Using the pattern for}%
\typeout{** the default language instead.}%
\else
\language=\csname l@#1\endcsname
\fi
#2}}
\providecommand{\BIBdecl}{\relax}
\BIBdecl

\bibitem{seligman2013navigating}
M.~E. Seligman, P.~Railton, R.~F. Baumeister, and C.~Sripada, ``Navigating into
  the future or driven by the past,'' \emph{Perspectives on psychological
  science}, vol.~8, no.~2, pp. 119--141, 2013.

\bibitem{finn2017deep}
C.~Finn and S.~Levine, ``Deep visual foresight for planning robot motion,'' in
  \emph{2017 IEEE International Conference on Robotics and Automation
  (ICRA)}.\hskip 1em plus 0.5em minus 0.4em\relax IEEE, 2017, pp. 2786--2793.

\bibitem{paxton2019visual}
C.~Paxton, Y.~Barnoy, K.~Katyal, R.~Arora, and G.~D. Hager, ``Visual robot task
  planning,'' in \emph{2019 international conference on robotics and automation
  (ICRA)}.\hskip 1em plus 0.5em minus 0.4em\relax IEEE, 2019, pp. 8832--8838.

\bibitem{hoque2020visuospatial}
R.~Hoque, D.~Seita, A.~Balakrishna, A.~Ganapathi, A.~K. Tanwani, N.~Jamali,
  K.~Yamane, S.~Iba, and K.~Goldberg, ``Visuospatial foresight for multi-step,
  multi-task fabric manipulation,'' \emph{arXiv preprint arXiv:2003.09044},
  2020.

\bibitem{xu2018neural}
D.~Xu, S.~Nair, Y.~Zhu, J.~Gao, A.~Garg, L.~Fei-Fei, and S.~Savarese, ``Neural
  task programming: Learning to generalize across hierarchical tasks,'' in
  \emph{2018 IEEE International Conference on Robotics and Automation
  (ICRA)}.\hskip 1em plus 0.5em minus 0.4em\relax IEEE, 2018, pp. 3795--3802.

\bibitem{huang2021dipn}
B.~Huang, S.~D. Han, A.~Boularias, and J.~Yu, ``Dipn: Deep interaction
  prediction network with application to clutter removal,'' in \emph{2021 IEEE
  International Conference on Robotics and Automation (ICRA)}.\hskip 1em plus
  0.5em minus 0.4em\relax IEEE, 2021, pp. 4694--4701.

\bibitem{huang2021visual}
B.~Huang, S.~D. Han, J.~Yu, and A.~Boularias, ``Visual foresight tree for
  object retrieval from clutter with nonprehensile rearrangement,'' \emph{arXiv
  preprint arXiv:2105.02857}, 2021.

\bibitem{suh2020surprising}
H.~Suh and R.~Tedrake, ``The surprising effectiveness of linear models for
  visual foresight in object pile manipulation,'' \emph{arXiv preprint
  arXiv:2002.09093}, 2020.

\bibitem{seita2021learning}
D.~Seita, P.~Florence, J.~Tompson, E.~Coumans, V.~Sindhwani, K.~Goldberg, and
  A.~Zeng, ``Learning to rearrange deformable cables, fabrics, and bags with
  goal-conditioned transporter networks,'' in \emph{2021 IEEE International
  Conference on Robotics and Automation (ICRA)}.\hskip 1em plus 0.5em minus
  0.4em\relax IEEE, 2021, pp. 4568--4575.

\bibitem{zeng2020transporter}
A.~Zeng, P.~Florence, J.~Tompson, S.~Welker, J.~Chien, M.~Attarian,
  T.~Armstrong, I.~Krasin, D.~Duong, V.~Sindhwani \emph{et~al.}, ``Transporter
  networks: Rearranging the visual world for robotic manipulation,''
  \emph{arXiv preprint arXiv:2010.14406}, 2020.

\bibitem{lim2021multi}
M.~H. Lim, A.~Zeng, B.~Ichter, M.~Bandari, E.~Coumans, C.~Tomlin, S.~Schaal,
  and A.~Faust, ``Multi-task learning with sequence-conditioned transporter
  networks,'' \emph{arXiv preprint arXiv:2109.07578}, 2021.

\bibitem{shridhar2022cliport}
M.~Shridhar, L.~Manuelli, and D.~Fox, ``Cliport: What and where pathways for
  robotic manipulation,'' in \emph{Conference on Robot Learning}.\hskip 1em
  plus 0.5em minus 0.4em\relax PMLR, 2022, pp. 894--906.

\bibitem{huang2022equivariant}
H.~Huang, D.~Wang, R.~Walter, and R.~Platt, ``Equivariant transporter
  network,'' \emph{arXiv preprint arXiv:2202.09400}, 2022.

\bibitem{ebert2017self}
F.~Ebert, C.~Finn, A.~X. Lee, and S.~Levine, ``Self-supervised visual planning
  with temporal skip connections.'' in \emph{CoRL}, 2017, pp. 344--356.

\bibitem{kossen2019structured}
J.~Kossen, K.~Stelzner, M.~Hussing, C.~Voelcker, and K.~Kersting, ``Structured
  object-aware physics prediction for video modeling and planning,''
  \emph{arXiv preprint arXiv:1910.02425}, 2019.

\bibitem{minderer2019unsupervised}
M.~Minderer, C.~Sun, R.~Villegas, F.~Cole, K.~Murphy, and H.~Lee,
  ``Unsupervised learning of object structure and dynamics from videos,''
  \emph{arXiv preprint arXiv:1906.07889}, 2019.

\bibitem{toussaint2015logic}
M.~Toussaint, ``Logic-geometric programming: An optimization-based approach to
  combined task and motion planning,'' in \emph{Twenty-Fourth International
  Joint Conference on Artificial Intelligence}, 2015.

\bibitem{garrett2021integrated}
C.~R. Garrett, R.~Chitnis, R.~Holladay, B.~Kim, T.~Silver, L.~P. Kaelbling, and
  T.~Lozano-P{\'e}rez, ``Integrated task and motion planning,'' \emph{Annual
  review of control, robotics, and autonomous systems}, vol.~4, pp. 265--293,
  2021.

\bibitem{kaelbling2010hierarchical}
L.~P. Kaelbling and T.~Lozano-P{\'e}rez, ``Hierarchical planning in the now,''
  in \emph{Workshops at the Twenty-Fourth AAAI Conference on Artificial
  Intelligence}, 2010.

\bibitem{song2020multi}
H.~Song, J.~A. Haustein, W.~Yuan, K.~Hang, M.~Y. Wang, D.~Kragic, and J.~A.
  Stork, ``Multi-object rearrangement with monte carlo tree search: A case
  study on planar nonprehensile sorting,'' in \emph{2020 IEEE/RSJ International
  Conference on Intelligent Robots and Systems (IROS)}.\hskip 1em plus 0.5em
  minus 0.4em\relax IEEE, 2020, pp. 9433--9440.

\bibitem{mukherjee2021reactive}
S.~Mukherjee, C.~Paxton, A.~Mousavian, A.~Fishman, M.~Likhachev, and D.~Fox,
  ``Reactive long horizon task execution via visual skill and precondition
  models,'' in \emph{2021 IEEE/RSJ International Conference on Intelligent
  Robots and Systems (IROS)}.\hskip 1em plus 0.5em minus 0.4em\relax IEEE,
  2021, pp. 5717--5724.

\bibitem{mcdonald2022guided}
M.~J. McDonald and D.~Hadfield-Menell, ``Guided imitation of task and motion
  planning,'' in \emph{Conference on Robot Learning}.\hskip 1em plus 0.5em
  minus 0.4em\relax PMLR, 2022, pp. 630--640.

\bibitem{qureshi2021nerp}
A.~H. Qureshi, A.~Mousavian, C.~Paxton, M.~C. Yip, and D.~Fox, ``Nerp: Neural
  rearrangement planning for unknown objects,'' \emph{arXiv preprint
  arXiv:2106.01352}, 2021.

\bibitem{labbe2020monte}
Y.~Labb{\'e}, S.~Zagoruyko, I.~Kalevatykh, I.~Laptev, J.~Carpentier, M.~Aubry,
  and J.~Sivic, ``Monte-carlo tree search for efficient visually guided
  rearrangement planning,'' \emph{IEEE Robotics and Automation Letters},
  vol.~5, no.~2, pp. 3715--3722, 2020.

\bibitem{hundt2020good}
A.~Hundt, B.~Killeen, N.~Greene, H.~Wu, H.~Kwon, C.~Paxton, and G.~D. Hager,
  ````good robot!'': Efficient reinforcement learning for multi-step visual
  tasks with sim to real transfer,'' \emph{IEEE Robotics and Automation
  Letters}, vol.~5, no.~4, pp. 6724--6731, 2020.

\bibitem{caruana1997multitask}
R.~Caruana, ``Multitask learning,'' \emph{Machine learning}, vol.~28, no.~1,
  pp. 41--75, 1997.

\bibitem{mulling2013learning}
K.~M{\"u}lling, J.~Kober, O.~Kroemer, and J.~Peters, ``Learning to select and
  generalize striking movements in robot table tennis,'' \emph{The
  International Journal of Robotics Research}, vol.~32, no.~3, pp. 263--279,
  2013.

\bibitem{kalashnikov2021mt}
D.~Kalashnikov, J.~Varley, Y.~Chebotar, B.~Swanson, R.~Jonschkowski, C.~Finn,
  S.~Levine, and K.~Hausman, ``Mt-opt: Continuous multi-task robotic
  reinforcement learning at scale,'' \emph{arXiv preprint arXiv:2104.08212},
  2021.

\bibitem{yang2020multi}
R.~Yang, H.~Xu, Y.~Wu, and X.~Wang, ``Multi-task reinforcement learning with
  soft modularization,'' \emph{arXiv preprint arXiv:2003.13661}, 2020.

\bibitem{hundt2021good}
A.~Hundt, A.~Murali, P.~Hubli, R.~Liu, N.~Gopalan, M.~Gombolay, and G.~D.
  Hager, ``" good robot! now watch this!": Repurposing reinforcement learning
  for task-to-task transfer,'' in \emph{5th Annual Conference on Robot
  Learning}, 2021.

\bibitem{finn2017one}
C.~Finn, T.~Yu, T.~Zhang, P.~Abbeel, and S.~Levine, ``One-shot visual imitation
  learning via meta-learning,'' in \emph{Conference on Robot Learning}.\hskip
  1em plus 0.5em minus 0.4em\relax PMLR, 2017, pp. 357--368.

\bibitem{duan2017one}
Y.~Duan, M.~Andrychowicz, B.~C. Stadie, J.~Ho, J.~Schneider, I.~Sutskever,
  P.~Abbeel, and W.~Zaremba, ``One-shot imitation learning,'' \emph{arXiv
  preprint arXiv:1703.07326}, 2017.

\bibitem{yu2018one}
T.~Yu, C.~Finn, A.~Xie, S.~Dasari, T.~Zhang, P.~Abbeel, and S.~Levine,
  ``One-shot imitation from observing humans via domain-adaptive
  meta-learning,'' \emph{arXiv preprint arXiv:1802.01557}, 2018.

\bibitem{long2015fully}
J.~Long, E.~Shelhamer, and T.~Darrell, ``Fully convolutional networks for
  semantic segmentation,'' in \emph{Proceedings of the IEEE conference on
  computer vision and pattern recognition}, 2015, pp. 3431--3440.

\bibitem{he2016deep}
K.~He, X.~Zhang, S.~Ren, and J.~Sun, ``Deep residual learning for image
  recognition,'' in \emph{Proceedings of the IEEE conference on computer vision
  and pattern recognition}, 2016, pp. 770--778.

\bibitem{diankov2010automated}
R.~Diankov, ``Automated construction of robotic manipulation programs,'' 2010.

\bibitem{chitta2012moveit}
S.~Chitta, I.~Sucan, and S.~Cousins, ``Moveit!'' \emph{IEEE Robotics \&
  Automation Magazine}, vol.~19, no.~1, pp. 18--19, 2012.

\bibitem{weiler2019general}
M.~Weiler and G.~Cesa, ``General e (2)-equivariant steerable cnns,''
  \emph{Advances in Neural Information Processing Systems}, vol.~32, 2019.

\bibitem{thomas2018tensor}
N.~Thomas, T.~Smidt, S.~Kearnes, L.~Yang, L.~Li, K.~Kohlhoff, and P.~Riley,
  ``Tensor field networks: Rotation-and translation-equivariant neural networks
  for 3d point clouds,'' \emph{arXiv preprint arXiv:1802.08219}, 2018.

\bibitem{blundell2015weight}
C.~Blundell, J.~Cornebise, K.~Kavukcuoglu, and D.~Wierstra, ``Weight
  uncertainty in neural network,'' in \emph{International conference on machine
  learning}.\hskip 1em plus 0.5em minus 0.4em\relax PMLR, 2015, pp. 1613--1622.

\end{thebibliography}

\newpage
\section*{SUPPLEMENTARY MATERIAL}
\label{sec: appendix}
The supplementary material is organized as follows:
\begin{itemize}
    \item Sec. \textit{A} discusses more details on the equivariance property of the dynamics of tabletop rearrangement.
    \item Sec. \textit{B} shows detailed results on both training and unseen tasks in simulation. It also contains ablation studies which compare more TVF variants.
    \item Sec. \textit{C} shows results of simulation experiments with more variable objects.
    \item Sec. \textit{D} describes more experiment details.
\end{itemize}

\subsection{SE(2) Equivariance of Dynamics}

\begin{figure}
    \centering
    \includegraphics[width=1\columnwidth]{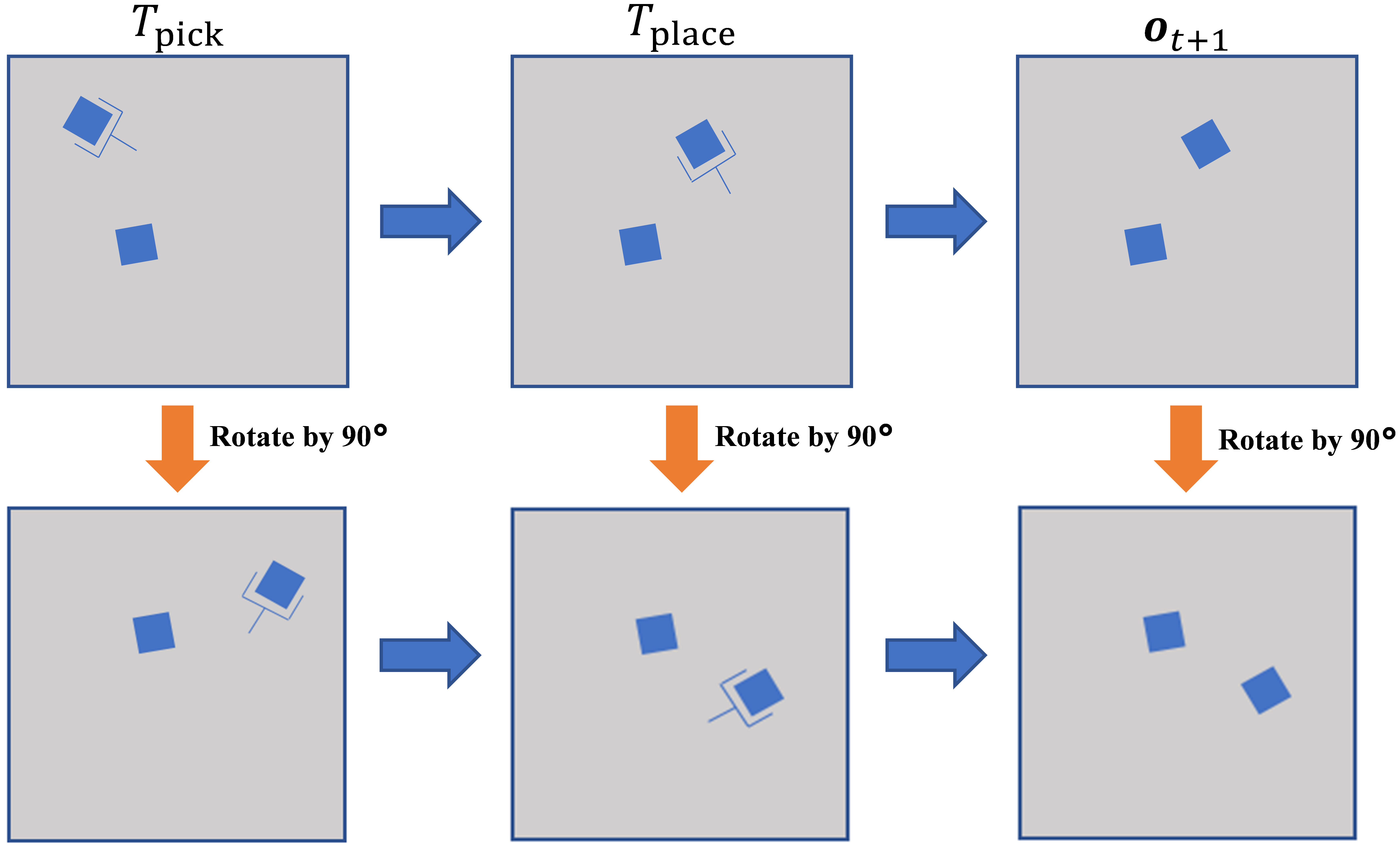}
    \caption{\textbf{Tabletop Rearrangment.} When the current observation and the pick-and-place action are transformed by $g\in \textit{SE}(2)$, the next-step observation will also be transformed by the same $g$.}
    \label{fig: equav}
    \vspace{-0.6cm}
\end{figure}

We assume a pre-defined 2D frame is attached to the infinite tabletop plane. 
All the coordinates and poses defined below are relative to this frame. 
Our observation is the orthographic top-down view $\mathbf{o}_t: \mathbb{R}^2\to \mathbb{R}^4$ of the whole tabletop workspace where $\mathbf{o}_t(u, v)$ gives the observed RGB and height value at position $\mathbf{p} = [u, v]^{T} \in \mathbb{R}^{2}$.
$g \in SE(2)$ can be parameterized with $g = (R(\theta), \mathbf{q})$ in which $\mathbf{q} = [\Delta u, \Delta v]^{T} \in \mathbb{R}^{2}$ represents the translation; $R(\theta)$ represents the rotation:
\begin{equation}
    R(\theta) = 
    \begin{bmatrix}
    \cos{\theta} & -\sin{\theta} \\
    \sin{\theta} & \cos{\theta}
    \end{bmatrix}
\end{equation}
The group action of $SE(2)$ on $\mathbf{p}\in \mathbb{R}^{2}$ and its inverse are defined respectively as:
\begin{equation}
    g \diamond \mathbf{p} \doteq R(\theta)\mathbf{p} + \mathbf{q} 
\end{equation}
\begin{equation}
    g^{-1} \diamond \mathbf{p} \doteq R(\theta)^{-1}\mathbf{p} - R( \theta)^{-1} \mathbf{q}
\end{equation}
We define the group action of $SE(2)$ on $\mathbf{o}_t$ as $g\cdot \mathbf{o}_t(\mathbf{p}) \doteq \mathbf{o}_t(g^{-1} \diamond \mathbf{p})$.
We denote $\mathbf{x}_t=(\mathbf{o}_t, \mathbf{a}_t)$ where $\mathbf{a}_t =({\textit{T}_{\rm{pick}}},\textit{T}_{\rm{place}}) \in SE(2) \times SE(2)$. 
The group action of ${SE}(2)$ on $\mathbf{a}_t$ is defined as $g \odot \mathbf{a}_{t} \doteq (g\circ \textit{T}_{\rm{pick}}, g\circ \textit{T}_{\rm{place}})$.
$\circ$ is the group operation of $SE(2)$ defined as $g_1\circ g_2 \doteq (R_1R_2, R_1\mathbf{q}_2+\mathbf{q}_1)$.
We then define the group action on $\mathbf{x}_{t}$ as $g \bullet \mathbf{x}_t \doteq (g\cdot \mathbf{o}_t, g \odot \mathbf{a}_t)$.
The $SE(2)$ equivariance property of the dynamics function $f: \mathbf{x}_t \to \mathbf{o}_{t+1}$ can be written as:
\begin{equation}
    f(g \bullet \mathbf{x}_t) = g \cdot f(\mathbf{x}_t)
    \label{eqn:equivariance:appendix}
\end{equation}
Intuitively, Eq.\ref{eqn:equivariance:appendix} describes the following property of the dynamics of tabletop rearrangement: if the current observation and the picking and placing poses are transformed by $g\in SE(2)$, the next-step observation should also be transformed by $g$ (Fig. \ref{fig: equav}). 
Our visual foresight (VF) model achieves translational equivariance by using a fully convolutional network (FCN) as the network architecture.
We leave the extension to $SE(2)$ equivariance as future work.
A promising direction is to represent the input (\textit{i.e.}, the observation and action) in a way such that it is compatible with an $SE(2)$ equivariant network architecture.

\subsection{Detailed Simulation Experiment Results}
In Tab. \ref{tab:results:traning_tasks} \& \ref{tab:results:novel_tasks}, we show full testing results on 6 training tasks and 8 unseen tasks. 
We show both the success rate and rate of progress for each task.
For the rate of progress, partial credit is also given to trials which are partially completed.
The rate of progress is defined as $\frac{\# \textnormal{of blocks in target poses}}{\# \textnormal{of blocks}}$.

TVF variants outperform GCTN in general, even with only one-step foresight (TVF-Small). 
The advantage of TVF variants over GCTN is more substantial on unseen tasks.
With the increase of demo number per task, the success rates of all methods grow in general. 
A substantial performance improvement is observed when the demo number per task increases from 1 to 10.
When the demo number increases further, the improvement is modest.
Given 100 and/or 1000 demos per task, the success rates for some tasks even decrease with the increase of demo number.
Similar results are also reported in \cite{seita2021learning}. 
With the increase of tree depth, TVF-Large outperforms TVF-Small in general.

Similar conclusions can also be drawn from Tab. \ref{tab:ablation1}-\ref{tab:ablation3} where we show more results on more TVF variants.
In these tables, we name each method with three letters ``K", ``M", and ``G", which represent the number of clusters in K-Means Clustering, the number of steps expanded by the multi-modal action proposal module, and the number of steps expanded by GCTN.
TVF-Small and -Large correspond to TVF-K2-M1-G0 and TVF-K3-M3-G0, respectively.
One additional observation from these tables is that the performance does not always improve with the increase of depth.
This counter-intuitive result will be explained in the next paragraph.
Another conclusion is that using more clusters for the action proposal improves performance in general.

The reason why in some cases the performance does not improve together with the increase of tree depth is threefold:
\begin{enumerate}
    \item If all the proposed actions are wrong at a given depth, the performance will not improve with the increase of tree depth.
    \item If the task can be finished within very a few steps, larger tree depth will not improve results.
    \item If the prediction of the dynamics model becomes unreliable as the tree grows, larger tree depth may deteriorate the performance.
\end{enumerate}
Therefore, to achieve better results with larger tree depth, the action proposal module and the dynamics model should improve simultaneously.
One future direction is to improve the generalization capability.

\begin{figure*}
    \centering
    \includegraphics[width=1.9\columnwidth]{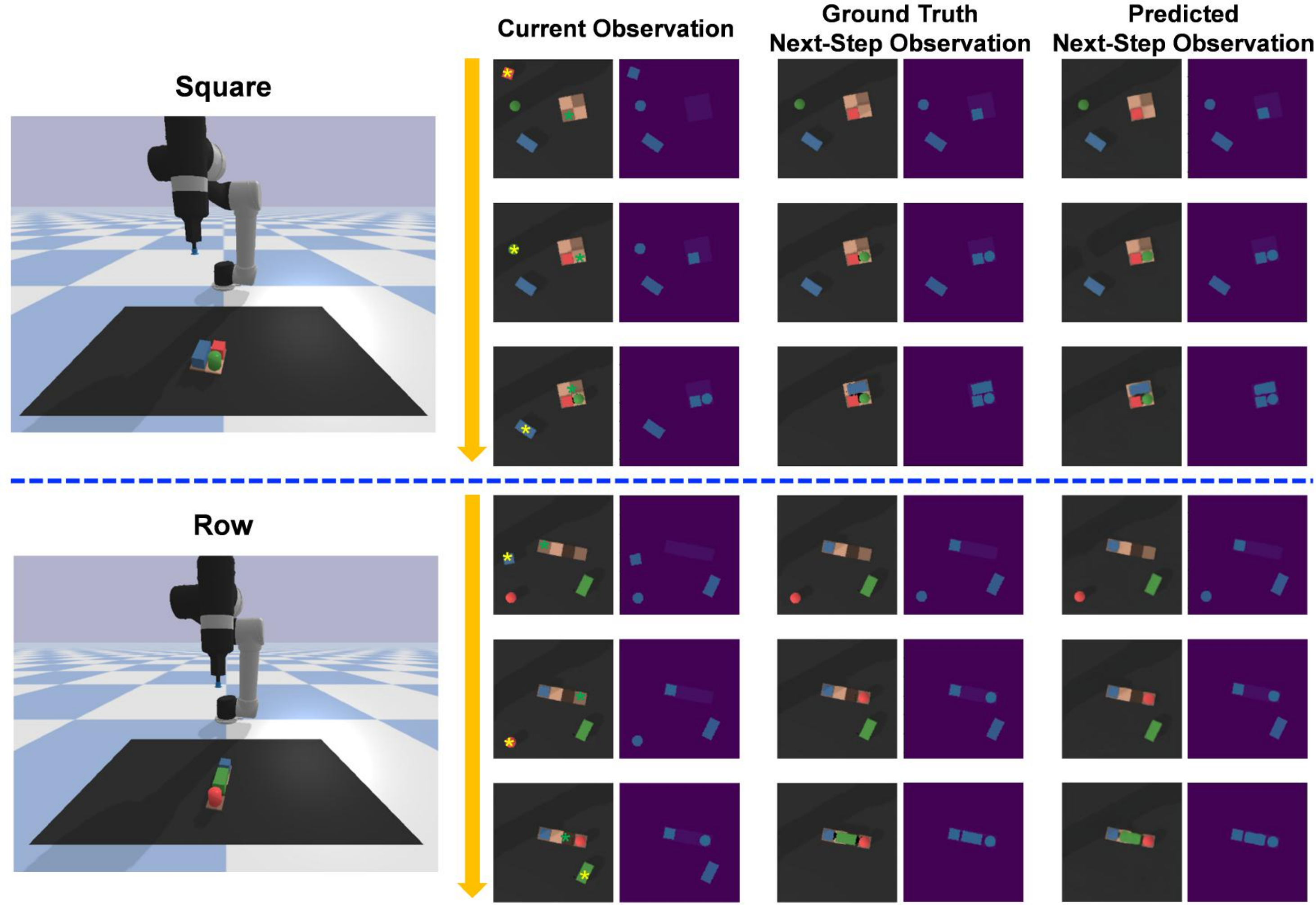}
    \caption{
    \textbf{VF Model Experiments on Variable Object Shapes and Colors.} 
    We show a rollout on the test data of each of the two tasks, Square and Row. 
    Current Observation shows the RGB image and heightmap of the current step. 
    The yellow and green stars indicate the picking position and placing position, respectively. 
    Ground Truth Next-Step Observation shows the ground truth next-step RGB image and heightmap. 
    Predicted Next-Step Observation shows the predicted next-step RGB image and heightmap of the next step. 
    The actions are expert actions.}
    \label{fig:rebuttal:variable}
\end{figure*}

\subsection{Experiments on Variable Objects}
We perform simulation experiments with more variable objects.
In particular, we experiment on two rearrangement tasks, Square and Row, containing a block, a cuboid, and a cylinder (Fig. \ref{fig:rebuttal:variable}).
The colors of the objects in the two tasks also vary.
In Square, the block, cuboid, and cylinder are painted red, blue, and green, respectively.
In Row, the block, cuboid, and cylinder are painted blue, green, and red, respectively.
10 demos per task are provided as the training data (20 demos in total).
Similar to the simulation experiments in \ref{subsec:experiments:results:unseen}, two random actions, which pick an object on the tabletop and place it at a collision-free pose, are also included in the collection of each expert demonstration.
We use this data to train the VF model and the GCTN for multi-modal action proposal.
Both random actions and oracle actions are used for training the VF model; only oracle actions are used for training the GCTN for the multi-modal action proposal.

We first evaluate our VF model on the test data of these two tasks and compare with the baseline method Latent Dynamics.
Qualitative results of our VF model are shown in Fig.~\ref{fig:rebuttal:variable}.
Quantitative results are shown in Tab.~\ref{tab:rebuttal:dynamics_result}.
Our VF model is able to retain the data efficiency and performance with variable object shapes and colors.
It outperforms Latent Dynamics by a large margin.
From Fig. \ref{fig:rebuttal:variable}, our VF model is able to predict accurate next-step RGB images and heightmaps given current observation and the pick-and-place action.
We have also observed that the color prediction of our VF model is worse than that tested on data that contains only red blocks in Sec.~\ref{subsec:experiments:results:dynamics} (L1 loss of 0.0296 compared to 0.0242 in Tab. \ref{tab:dynamics_result}; lower is better).
This is expected because there are more colors in this experiment, making color prediction more challenging.
The height prediction is better than that tested on data with only red blocks (L1 loss of 0.0100 compared to 0.0136 in Tab. \ref{tab:dynamics_result}).
This is also within expectation because there are only two tasks in this experiment while there are six tasks in Sec. \ref{subsec:experiments:results:unseen}.

We also test the TVF-Small variant on the test data of these two tasks.
We compare with GCTN.
Results are shown in Tab. \ref{tab:rebuttal:task_result}.
TVF-Small outperforms GCTN on both tasks.
In both tasks, the success rates of TVF-Small are higher than 80\%.
Our task planning method is able to retain the data efficiency and performance on tasks with variable objects and colors.

\begin{table}[t]
    \setlength\tabcolsep{10pt}
    \caption{
    \textbf{Visual Foresight Prediction Results of Experiments on Variable Object Shapes and Colors.} 
    The table shows the visual foresight prediction results of testing our VF model and Latent Dynamics on the test data with variable objects and colors. 
    Both methods are trained with 10 demos per training task (20 demos per task).
    The table shows the L1 loss of the RGB channels and the height channel between the predicted observation and the ground truth observation.
    The images are normalized.
    The actions are expert actions.
    Lower is better.}
    \centering
    \begin{tabular}{@{}lccccc@{}}
    \toprule
        
        Method & Color & Height\\
        \midrule
        Latent Dynamics   & 0.1082          & 0.0935\\
        Ours              & \textbf{0.0296} & \textbf{0.0100}\\
    \bottomrule
    \end{tabular}
    \label{tab:rebuttal:dynamics_result}
\end{table}

\begin{table}[t]
    \setlength\tabcolsep{10pt}
    \caption{
    \textbf{Success Rates of Experiments on Variable Object Shapes and Colors.} 
    The table shows the success rates (\%) of testing TVF-Small and GCTN on two tasks with variable objects and colors.
    Each task is tested on 20 rollouts.}
    \centering
    \begin{tabular}{@{}lccccc@{}}
    \toprule
        
        Method & Square & Row\\
        \midrule
        GCTN             & 81.7          & 68.3\\
        TVF-Small (Ours) & \textbf{85.0} & \textbf{81.7}\\
    \bottomrule
    \end{tabular}
    \label{tab:rebuttal:task_result}
\end{table}

\begin{figure}[h!]
    \centering
    \includegraphics[width=\columnwidth]{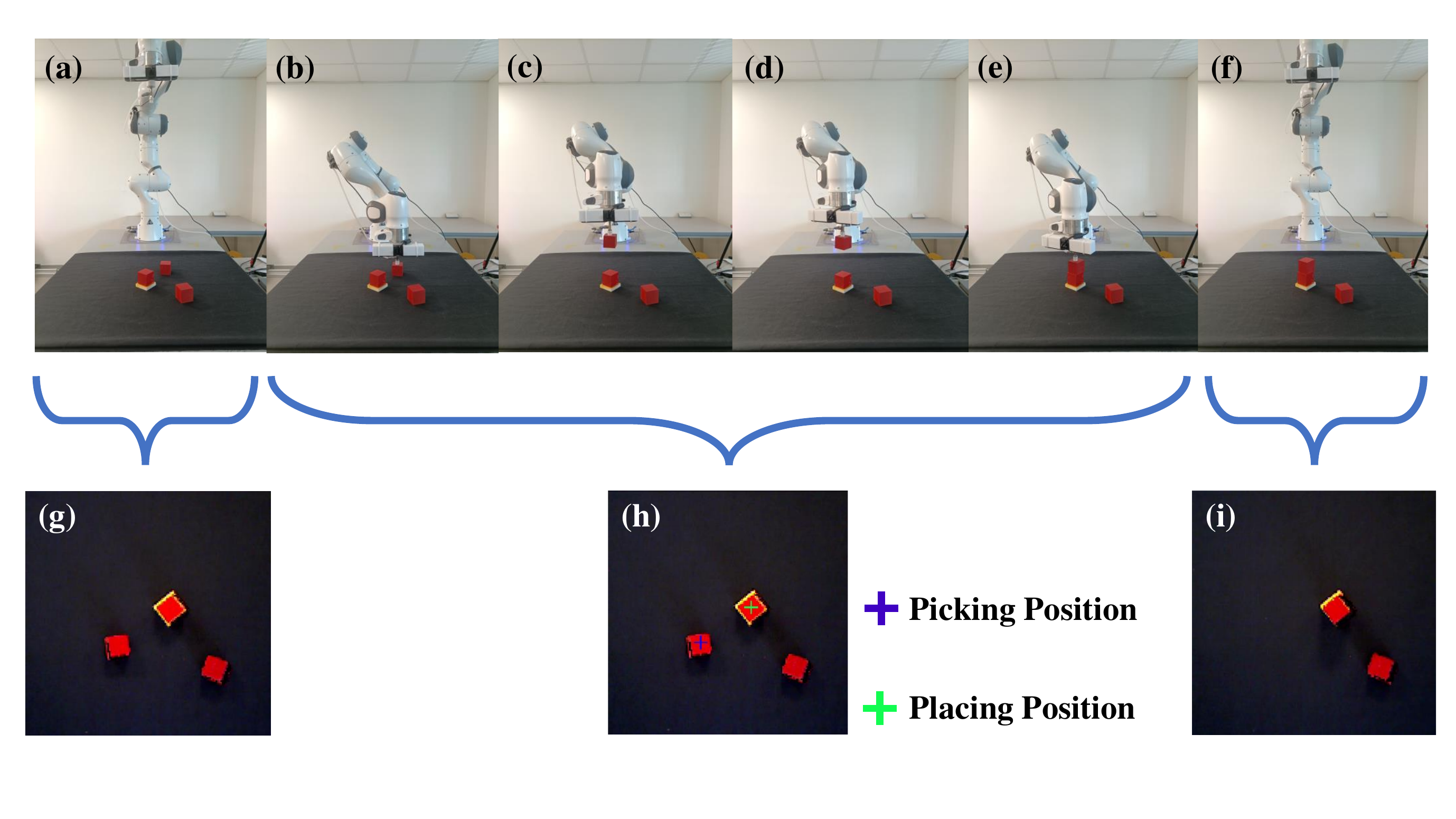}
    \caption{\textbf{Real Robot Data Collection of a Step.}
    (a) In each step of a rollout, the robot arm moves to the top of the workspace and captures the top-down view RGB image and heightmap.
    (g) shows the captured top-down RGB image.
    The top-down image will then show up on the computer and a human expert clicks on two points on the image to specify the picking and placing positions, respectively.
    (h) shows the clicked points.
    (b) The robot then moves to the picking position and picks up the block.
    (c) It then moves towards the placing position.
    (d) Before placing, the human expert manually specifies the placing rotation angle.
    (e) Finally, the robot places the block down at the placing position.
    (f) The robot moves to capture the top-down view and start a new step.
    If the task is completed, this view will be saved as the goal.
    (i) shows the top-down RGB image of the observation in (f).
    The process is repeated until the task is completed.}
    \label{fig:rebuttal:step}
\end{figure}

\subsection{More Experiment Details}
\begin{figure}
    \centering
    \includegraphics[width=\columnwidth]{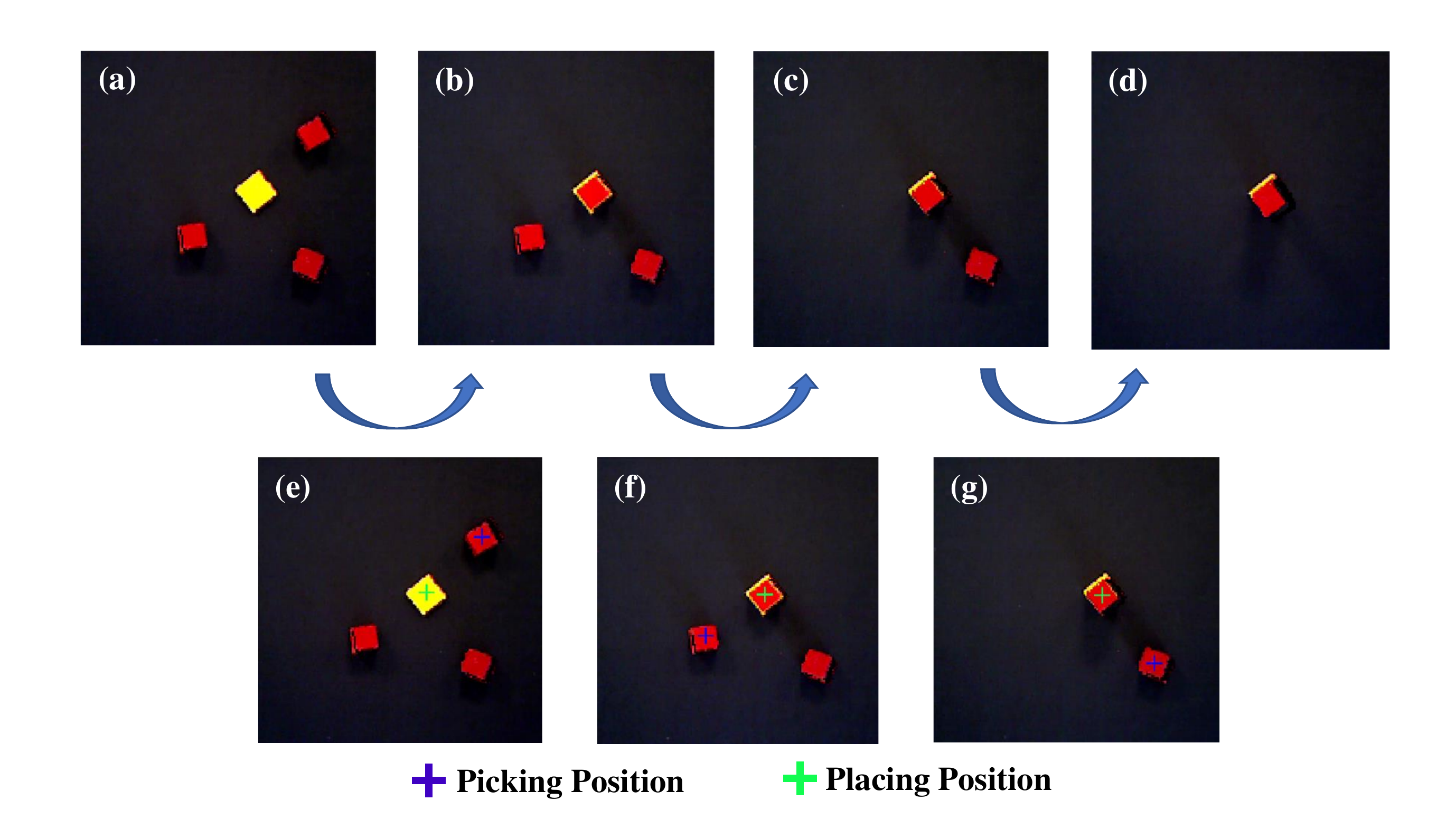}
    \caption{\textbf{Real Robot Data Collection of a Rollout.}
    (a), (b), (c), and (d) show the top-down RGB images from the initial to the end of a rollout.
    The task is Tower.
    (e), (f), and (g) show the picking and placing positions of the three steps specified by the human expert.
    Random actions are not shown.}
    \vspace{-0.5cm}
    \label{fig:rebuttal:rollout}
\end{figure}

To collect training data in real robot experiments (Sec. \ref{subsec:experiments:real}), we implement an efficient way for a human expert to teleoperate the robot to pick and place blocks.
See Fig. \ref{fig:rebuttal:step} and its caption for a detailed description of data collection of a step.
Fig. \ref{fig:rebuttal:rollout} shows the data collection of a rollout.
The top-down RGB image and heightmap, the picking and placing positions, and the placing rotation angle of each step are collected during a rollout.
The top-down RGB image and heightmap at the end of the rollout are also collected as the goal.
To increase data variability, two random actions are also collected in each rollout similar to the simulation experiments in Sec. \ref{subsec:experiments:results:unseen}.
The human expert teleoperates the robot to pick a block and place it at a random pose which is collision-free.
We perform a background filtering for both RGB images and heighmaps during data collection and testing.
Specifically, we convert RGB to HSV and set thresholds on the height and V value.
For each filtered pixel, we set the RGB and height value as zero.
We find that GCTN is not able to learn well on real data without the background filtering.

Tab. \ref{tab:hyperparameter} shows the hyperparameters we use for training our VF model, GCTN, and Latent Dynamics in the paper.
More details can be found in our project website: \url{https://chirikjianlab.github.io/tvf/}

\begin{table}[t]
  \setlength\tabcolsep{5.0pt}
  \caption{
  \textbf{Hyperparameters}
  }
  \centering
  \footnotesize
  \begin{tabular}{@{}lcccccccccccccccc@{}} % centering
  \toprule
 Hyperparameter & Value 
 \\
 \midrule
 Learning Rate (VF)  & $1\times10^{-4}$
 \\
 Minibatch Size (VF) & 1
 \\
 Training Steps (VF) & $6\times 10^{4}$
 \\
 Learning Rate (Latent Dynamics)  & $1\times10^{-4}$
 \\
 Minibatch Size (Latent Dynamics) & 1
 \\
 Training Steps (Latent Dynamics) & $6\times 10^{4}$
 \\
 Tree Value Coefficient $C$ (TVF) & 1
 \\
 Discount Factor $\gamma$ (TVF) & 0.99
 \\
 K-Means Clustering Threshold Coefficient $\alpha$ (TVF) & 0.01
 \\
 Top N number in K-Means Clustering $N$ (TVF) & 100
 \\
 Number of Rotation Bin for GCTN $R$ (GCTN) & 36
 \\
  \bottomrule \vspace{-0.8em} \\
  \end{tabular}
  \label{tab:hyperparameter}
\end{table}

\begin{table*}[t]
  \setlength\tabcolsep{10.5pt}
  \caption{
  \textbf{Simulation Experiment Results on Training Tasks.}
  We show the average success rate (\%) / rate of progress (\%) on the test data of each training task v.s. \# of demonstrations (1, 10, 100, or 1000) per task in the training data.
  Higher is better.
  }
  \centering
  \footnotesize
  \begin{tabular}{@{}lcccccccccccccccc@{}} % centering
  \toprule
  & \multicolumn{4}{c}{Row} & \multicolumn{4}{c}{Square}\\
 \cmidrule(lr){2-5} \cmidrule(lr){6-9}
 Method & 1 & 10 & 100 & 1000 & 1 & 10 & 100 & 1000\\
 \midrule
 GCTN      &           8.3/35.0          &           98.3/99.4           & \textbf{95.0}/\textbf{98.3} & \textbf{100.0}/\textbf{100.0}   &          0.0/34.2          &           93.3/96.7           &          65.0/84.2          &           93.3/96.7          \\
 TVF-Small &          11.7/\textbf{42.2} & \textbf{100.0}/\textbf{100.0} & \textbf{95.0}/\textbf{98.3} & \textbf{100.0}/\textbf{100.0}   &          1.7/37.1          &           90.0/95.0           &          80.0/90.8          &           98.3/99.2          \\
 TVF-Large & \textbf{15.0}/\textbf{42.2} & \textbf{100.0}/\textbf{100.0} & \textbf{95.0}/\textbf{98.3} & \textbf{100.0}/\textbf{100.0}   & \textbf{3.3}/\textbf{40.0} & \textbf{100.0}/\textbf{100.0} & \textbf{90.0}/\textbf{97.1} & \textbf{100.0}/\textbf{100.0}\\
 \midrule
 & \multicolumn{4}{c}{T-shape} & \multicolumn{4}{c}{Tower}\\
 \cmidrule(lr){2-5} \cmidrule(lr){6-9}
 Method & 1 & 10 & 100 & 1000 & 1 & 10 & 100 & 1000\\
 \midrule
 GCTN      &          1.7/34.2          &             80.0/93.3          &          95.0/98.7          &          95.0/\textbf{98.7} &          3.3/32.8          & \textbf{100.0}/\textbf{100.0} &           98.3/98.3           & \textbf{100.0}/\textbf{100.0}\\
 TVF-Small & \textbf{3.3}/36.2          &             90.0/95.8          & \textbf{96.7}/\textbf{99.2} &          95.0/97.5          & \textbf{5.0}/\textbf{42.8} & \textbf{100.0}/\textbf{100.0} & \textbf{100.0}/\textbf{100.0} & \textbf{100.0}/\textbf{100.0}\\
 TVF-Large &          1.7/\textbf{37.1} &    \textbf{96.7}/\textbf{98.7} & \textbf{96.7}/\textbf{99.2} & \textbf{96.7}/\textbf{98.7} & \textbf{5.0}/\textbf{42.8} & \textbf{100.0}/\textbf{100.0} & \textbf{100.0}/\textbf{100.0} &                     98.3/98.9\\
\midrule
 & \multicolumn{4}{c}{Pyramid} & \multicolumn{4}{c}{Palace}\\
 \cmidrule(lr){2-5} \cmidrule(lr){6-9}
 Method & 1 & 10 & 100 & 1000 & 1 & 10 & 100 & 1000\\
 \midrule
 GCTN      &          0.0/31.4          &          73.3/93.1          &          83.3/\textbf{96.7} & \textbf{81.7}/\textbf{95.6}   & 0.0/32.4          &          61.7/88.8          &          78.3/95.2          & \textbf{85.0}/\textbf{96.4}\\
 TVF-Small & \textbf{1.7}/\textbf{34.7} &          75.0/\textbf{93.3} & \textbf{85.0}/96.1          &          61.7/88.1            & 0.0/\textbf{36.9} & \textbf{75.0}/91.9          &          80.0/95.7          &          80.0/96.0         \\
 TVF-Large &          0.0/34.2          & \textbf{80.0}/92.8          &          81.7/95.6          &          66.7/89.2            & 0.0/33.8          &          71.7/\textbf{92.6} & \textbf{85.0}/\textbf{96.9} &          83.3/95.5         \\
  \bottomrule \\
  \end{tabular}
  \label{tab:results:traning_tasks}
\end{table*}

\begin{table*}[!h]
  \setlength\tabcolsep{10.5pt}
  \caption{
  \textbf{Simulation Experiment Results on Unseen Tasks.}
  We show the average success rate (\%) / rate of progress (\%) on the test data of each unseen task v.s. \# of demonstrations (1, 10, 100, or 1000) per task in the training data.
  Higher is better.
  }
  \centering
  \footnotesize
  \begin{tabular}{@{}lcccccccccccccccc@{}} % centering
  \toprule
 & \multicolumn{4}{c}{Plane Square} & \multicolumn{4}{c}{Plane T} \\
 \cmidrule(lr){2-5} \cmidrule(lr){6-9}
 Method & 1 & 10 & 100 & 1000 & 1 & 10 & 100 & 1000\\
 \midrule
  GCTN      &          1.7/43.8          &           86.7/96.3           &          95.0/98.7          &           96.7/98.7             &           5.0/39.4          &          78.3/92.8          &           93.3/97.8           &          90.0/96.7         \\
  TVF-Small &          3.3/\textbf{45.0} &           98.3/99.6           & \textbf{96.7}/\textbf{99.2} & \textbf{100.0}/\textbf{100.0}   &           3.3/43.3          & \textbf{90.0}/\textbf{96.7} & \textbf{100.0}/\textbf{100.0} & \textbf{98.3}/\textbf{99.4}\\
  TVF-Large & \textbf{5.0}/\textbf{45.0} & \textbf{100.0}/\textbf{100.0} & \textbf{96.7}/\textbf{99.2} &           98.3/99.2             & \textbf{15.0}/\textbf{46.1} &          86.7/95.6          & \textbf{100.0}/\textbf{100.0} &          95.0/98.3         \\
  \midrule
  & \multicolumn{4}{c}{Stair 2} & \multicolumn{4}{c}{Twin Tower}\\
 \cmidrule(lr){2-5} \cmidrule(lr){6-9} 
 Method & 1 & 10 & 100 & 1000 & 1 & 10 & 100 & 1000\\
 \midrule
  GCTN      &          3.3/38.3          &          85.0/95.0          &          46.7/82.2          &           68.3/89.4           & 0.0/25.8          &          88.3/94.7          &          55.0/71.9          &          85.0/95.8         \\
  TVF-Small & \textbf{6.7}/\textbf{43.9} & \textbf{98.3}/\textbf{99.4} &          71.7/90.6          &           90.0/96.7           & 0.0/\textbf{34.2} & \textbf{98.3}/\textbf{98.9} & \textbf{85.0}/90.0          & \textbf{93.3}/\textbf{97.5}\\
  TVF-Large &          3.3/40.0          &          96.7/97.8          & \textbf{95.0}/\textbf{97.8} & \textbf{100.0}/\textbf{100.0} & 0.0/32.5          &          96.7/98.1          & \textbf{85.0}/\textbf{91.7} &          91.7/96.1         \\
 \midrule
 & \multicolumn{4}{c}{Stair 3} & \multicolumn{4}{c}{Building}\\
 \cmidrule(lr){2-5} \cmidrule(lr){6-9} 
 Method & 1 & 10 & 100 & 1000 & 1 & 10 & 100 & 1000\\
 \midrule
  GCTN      & 0.0/30.6          &          45.0/86.4          &          23.3/67.5          &          16.7/76.9            & 0.0/26.3          &           5.0/55.3          &           0.0/57.0          &           3.3/54.3         \\
  TVF-Small & 0.0/\textbf{38.6} &          63.3/91.1          &          33.3/75.3          &          46.7/85.0            & 0.0/\textbf{30.3} &           8.3/\textbf{58.7} & \textbf{10.0}/\textbf{66.3} &          11.7/64.7         \\
  TVF-Large & 0.0/32.8          & \textbf{81.7}/\textbf{94.7} & \textbf{56.7}/\textbf{86.9} & \textbf{90.0}/\textbf{97.2}   & 0.0/26.3          & \textbf{13.3}/58.0          &           6.7/60.0          & \textbf{25.0}/\textbf{68.3}\\
 \midrule
  & \multicolumn{4}{c}{Pallet} & \multicolumn{4}{c}{Rectangle}\\
 \cmidrule(lr){2-5} \cmidrule(lr){6-9} \cmidrule(lr){10-13}
 Method & 1 & 10 & 100 & 1000 & 1 & 10 & 100 & 1000\\
 \midrule
  GCTN      & 0.0/31.5          &          23.3/82.5          &          51.7/78.5          &          31.7/84.0          &          0.0/31.1          &          31.7/84.7          &          26.7/68.3          &          41.7/79.4         \\
  TVF-Small & 0.0/\textbf{34.2} &          60.0/91.5          &          61.7/83.5          &          65.0/94.0          &          0.0/\textbf{35.3} &          55.0/88.1          &          40.0/77.2          &          75.0/89.7         \\
  TVF-Large & 0.0/32.1          & \textbf{75.0}/\textbf{94.4} & \textbf{70.0}/\textbf{91.7} & \textbf{90.0}/\textbf{97.5} &          0.0/30.8          & \textbf{78.3}/\textbf{94.2} & \textbf{63.3}/\textbf{86.4} & \textbf{95.0}/\textbf{98.6}\\
  \bottomrule\\
  \end{tabular}
  \label{tab:results:novel_tasks}
\end{table*}

\cleardoublepage

\begin{table*}[!t]
  \setlength\tabcolsep{13pt}
  \caption{
  \textbf{Ablation Study (10 Demos).}
  We show the average success rate (\%) on the test data of unseen tasks.
  The number of demonstrations per task in the training data is 10.
  Higher is better.
  }
  \centering
  \footnotesize
  \begin{tabular}{@{}lcccccccccccccccc@{}} % centering
  \toprule
 Method & Plane Square & Plane T & Stair 2 & Twin Tower & Stair 3 & Building & Pallet & Rectangle\\
 \midrule
 TVF-K2-M1-G0  &  98.3 & 90.0 & 98.3 & 98.3 & 63.3 & 8.3 & 60.0 & 55.0\\
 TVF-K2-M2-G0  & 100.0 & 86.7 & 93.3 & 98.3 & 70.0 & 3.3 & 65.0 & 55.0\\
 TVF-K2-M3-G0  & 100.0 & 85.0 & 93.3 & 95.0 & 70.0 & 8.3 & 56.7 & 55.0\\
 TVF-K2-M4-G0  & 100.0 & 85.0 & 93.3 & 95.0 & 63.3 & 6.7 & 68.3 & 61.7\\
 TVF-K2-M4-G1  & 100.0 & 85.0 & 93.3 & 98.3 & 65.0 & 8.3 & 66.7 & 58.3\\
 \midrule
 TVF-K3-M1-G0  & 100.0 & 88.3 & 100.0 & 98.3 & 68.3 &  3.3 & 76.7 & 76.7\\
 TVF-K3-M2-G0  & 100.0 & 88.3 &  96.7 & 90.0 & 76.7 &  3.3 & 66.7 & 76.7\\
 TVF-K3-M3-G0  & 100.0 & 86.7 &  96.7 & 96.7 & 81.7 & 13.3 & 75.0 & 78.3\\
 TVF-K3-M4-G0  & 100.0 & 86.7 &  96.7 & 95.0 & 71.7 & 10.0 & 70.0 & 86.7\\
 TVF-K3-M4-G1  & 100.0 & 86.7 &  96.7 & 95.0 & 76.7 & 13.3 & 68.3 & 86.7\\
  \bottomrule \\
  \end{tabular}
  \label{tab:ablation1}
\end{table*}

\begin{table*}[!h]
  \setlength\tabcolsep{13pt}
  \caption{
  \textbf{Ablation Study (100 Demos).}
  We show the average success rate (\%) on the test data of unseen tasks.
  The number of demonstrations per task in the training data is 100.
  Higher is better.
  }
  \centering
  \footnotesize
  \begin{tabular}{@{}lcccccccccccccccc@{}} % centering
  \toprule
 Method & Plane Square & Plane T & Stair 2 & Twin Tower & Stair 3 & Building & Pallet & Rectangle\\
 \midrule
 TVF-K2-M1-G0  & 96.7 & 100.0 & 71.7 & 85.0 & 33.3 & 10.0 & 61.7 & 40.0\\
 TVF-K2-M2-G0  & 96.7 & 100.0 & 75.0 & 80.0 & 38.3 &  1.7 & 60.0 & 43.3\\
 TVF-K2-M3-G0  & 96.7 & 100.0 & 85.0 & 80.0 & 43.3 &  1.7 & 60.0 & 51.7\\
 TVF-K2-M4-G0  & 96.7 & 100.0 & 85.0 & 83.3 & 41.7 &  8.3 & 63.3 & 51.7\\
 TVF-K2-M4-G1  & 96.7 & 100.0 & 85.0 & 88.3 & 43.3 &  5.0 & 58.3 & 51.7\\
 \midrule
 TVF-K3-M1-G0  & 95.0 & 100.0 & 80.0 & 88.3 & 51.7 & 10.0 & 63.3 & 53.3\\
 TVF-K3-M2-G0  & 96.7 & 100.0 & 91.7 & 95.0 & 55.0 &  1.7 & 68.3 & 60.0\\
 TVF-K3-M3-G0  & 96.7 & 100.0 & 95.0 & 85.0 & 56.7 &  6.7 & 70.0 & 63.3\\
 TVF-K3-M4-G0  & 96.7 & 100.0 & 93.3 & 88.3 & 53.3 & 10.0 & 63.3 & 68.3\\
 TVF-K3-M4-G1  & 96.7 & 100.0 & 93.3 & 90.0 & 51.7 & 11.7 & 63.3 & 66.7\\
  \bottomrule \\
  \end{tabular}
  \label{tab:ablation2}
\end{table*}

\begin{table*}[!h]
  \setlength\tabcolsep{13pt}
  \caption{
  \textbf{Ablation Study (1000 Demos).}
  We show the average success rate (\%) on the test data of unseen tasks.
  The number of demonstrations per task in the training data is 1000.
  Higher is better.
  }
  \centering
  \footnotesize
  \begin{tabular}{@{}lcccccccccccccccc@{}} % centering
  \toprule
 Method & Plane Square & Plane T & Stair 2 & Twin Tower & Stair 3 & Building & Pallet & Rectangle\\
 \midrule
 TVF-K2-M1-G0  & 100.0 & 98.3 & 90.0 & 93.3 & 46.7 & 11.7 & 65.0 & 75.0\\
 TVF-K2-M2-G0  &  98.3 & 98.3 & 90.0 & 96.7 & 58.3 &  5.0 & 76.7 & 80.0\\
 TVF-K2-M3-G0  & 100.0 & 98.3 & 91.7 & 98.3 & 50.0 & 10.0 & 75.0 & 88.3\\
 TVF-K2-M4-G0  & 100.0 & 98.3 & 91.7 & 98.3 & 61.7 & 26.7 & 71.7 & 88.3\\
 TVF-K2-M4-G1  & 100.0 & 98.3 & 91.7 & 96.7 & 56.7 & 33.3 & 80.0 & 95.0\\
 \midrule
 TVF-K3-M1-G0  & 100.0 & 96.7 & 100.0 & 98.3 & 78.3 & 15.0 & 88.3 & 83.3\\
 TVF-K3-M2-G0  &  98.3 & 95.0 & 100.0 & 98.3 & 85.0 & 13.3 & 93.3 & 93.3\\
 TVF-K3-M3-G0  &  98.3 & 95.0 & 100.0 & 91.7 & 90.0 & 25.0 & 90.0 & 95.0\\
 TVF-K3-M4-G0  & 100.0 & 95.0 &  98.3 & 95.0 & 91.7 & 36.7 & 88.3 & 86.7\\
 TVF-K3-M4-G1  & 100.0 & 95.0 &  98.3 & 98.3 & 83.3 & 40.0 & 86.7 & 86.7\\
  \bottomrule\\
  \end{tabular}
  \label{tab:ablation3}
\end{table*}

%                                   % the textheight of the last page by a suitable amount.
%                                   % This command does not take effect until the next page
%                                   % so it should come on the page before the last. Make
%                                   % sure that you do not shorten the textheight too much.

\end{document}